\documentclass{article}

\usepackage{microtype}
\usepackage{graphicx}
\usepackage{booktabs} % professional-quality tables

\usepackage{makecell}
\usepackage{soul}
\setul{0.4ex}{0.1ex}
\usepackage{array}
\usepackage{longtable}
\usepackage{url}
\usepackage{hyperref}
\usepackage{enumitem}
\usepackage{caption}
\usepackage{float}
\usepackage{dblfloatfix}

\usepackage[accepted]{icml2025}

\usepackage{amsmath}
\usepackage{amssymb}
\usepackage{mathtools}

\makeatletter
\gdef\isaccepted{1}
\renewcommand{\Notice@String}{}

\begin{document}

\twocolumn[
\icmltitle{\smash{\makebox[0pt][r]{\raisebox{-1.4em}{\includegraphics[height=3em]{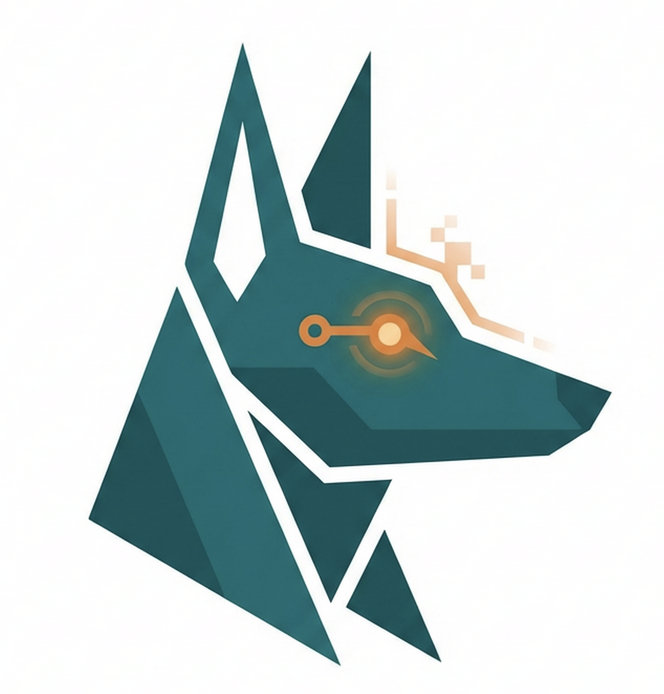}}\hspace{2.0em}}}Agentic Jackal: Live Execution and \\Semantic Value Grounding for Text-to-JQL}
\begin{icmlauthorlist}
\icmlauthor{Vishnu Murali\textsuperscript{*}}{}
\icmlauthor{Anmol Gulati}{}
\icmlauthor{Elias Lumer}{}
\icmlauthor{Kevin Frank}{}
\icmlauthor{Sindy Campagna}{}
\icmlauthor{Vamse Kumar Subbiah}{}
\\
\textit{Commercial Technology and Innovation Office, PricewaterhouseCoopers, U.S.}
% \\
% \texttt{vishnu.v.murali@pwc.com}
\end{icmlauthorlist}

% \icmlaffiliation{}{Commercial Technology and Innovation Office, PricewaterhouseCoopers, U.S.}
\icmlcorrespondingauthor{Vishnu Murali}{vishnu.v.murali@pwc.com}

\vskip 0.3in
]
% \printAffiliationsAndNotice{}
\setcounter{footnote}{1}
\renewcommand{\thefootnote}{\fnsymbol{footnote}}
\footnotetext{Correspondence to: vishnu.v.murali@pwc.com}

\begin{abstract}
Translating natural language into Jira Query Language (JQL) requires
resolving ambiguous field references, instance-specific categorical
values, and complex Boolean predicates.
Single-pass LLMs cannot discover which categorical values
(e.g., component names or fix versions) actually exist in a given
Jira instance, nor can they verify generated queries against a live
data source, limiting accuracy on paraphrased or ambiguous requests.
No open, execution-based benchmark exists for mapping natural
language to JQL.
We introduce Jackal, the first large-scale, execution-based
text-to-JQL benchmark comprising 100,000 validated NL--JQL pairs
on a live Jira instance with over 200,000 issues.
To establish baselines on Jackal, we propose Agentic Jackal, a
tool-augmented agent that equips LLMs with live query execution
via the Jira MCP server and JiraAnchor, a semantic retrieval tool
that resolves natural-language mentions of categorical values through
embedding-based similarity search.
Among 9 frontier LLMs evaluated, single-pass models average only
43.4\% execution accuracy on short natural-language queries,
highlighting that text-to-JQL remains an open challenge.
The agentic approach improves 7 of 9 models, with a 9.0\% relative
gain on the most linguistically challenging variant; in a controlled
ablation isolating JiraAnchor, categorical-value accuracy rises from
48.7\% to 71.7\%, with component-field accuracy jumping from
16.9\% to 66.2\%.
Our analysis identifies inherent semantic ambiguities, such as
issue-type disambiguation and text-field selection, as the dominant
failure modes rather than value-resolution errors, pointing to
concrete directions for future work.
We publicly release the benchmark, all agent transcripts, and
evaluation code to support reproducibility.
\end{abstract}

% ===========================================================================
% SECTION 1: INTRODUCTION
% ===========================================================================
\section{Introduction}

Natural language interfaces that map user requests into domain-specific query languages have become increasingly effective with the rise of Large Language Models (LLMs), reducing the learning curve of formal syntax for non-expert users. Atlassian Jira, one of the most widely used issue
tracking and project management platforms, exemplifies the demand for
such interfaces. Its Jira Query Language (JQL) supports complex
Boolean logic, custom fields, and temporal predicates, but presents
significant barriers for its 300,000+
users~\citep{atlassian2025jql}. JQL further compounds the challenge
with project- and instance-specific custom fields, linked-issue
traversals, and value vocabularies that vary across deployments,
making execution-based evaluation against a live instance
essential~\citep{lei2025spider}.

\begin{figure*}[t]
\centering
\includegraphics[width=0.75\textwidth]{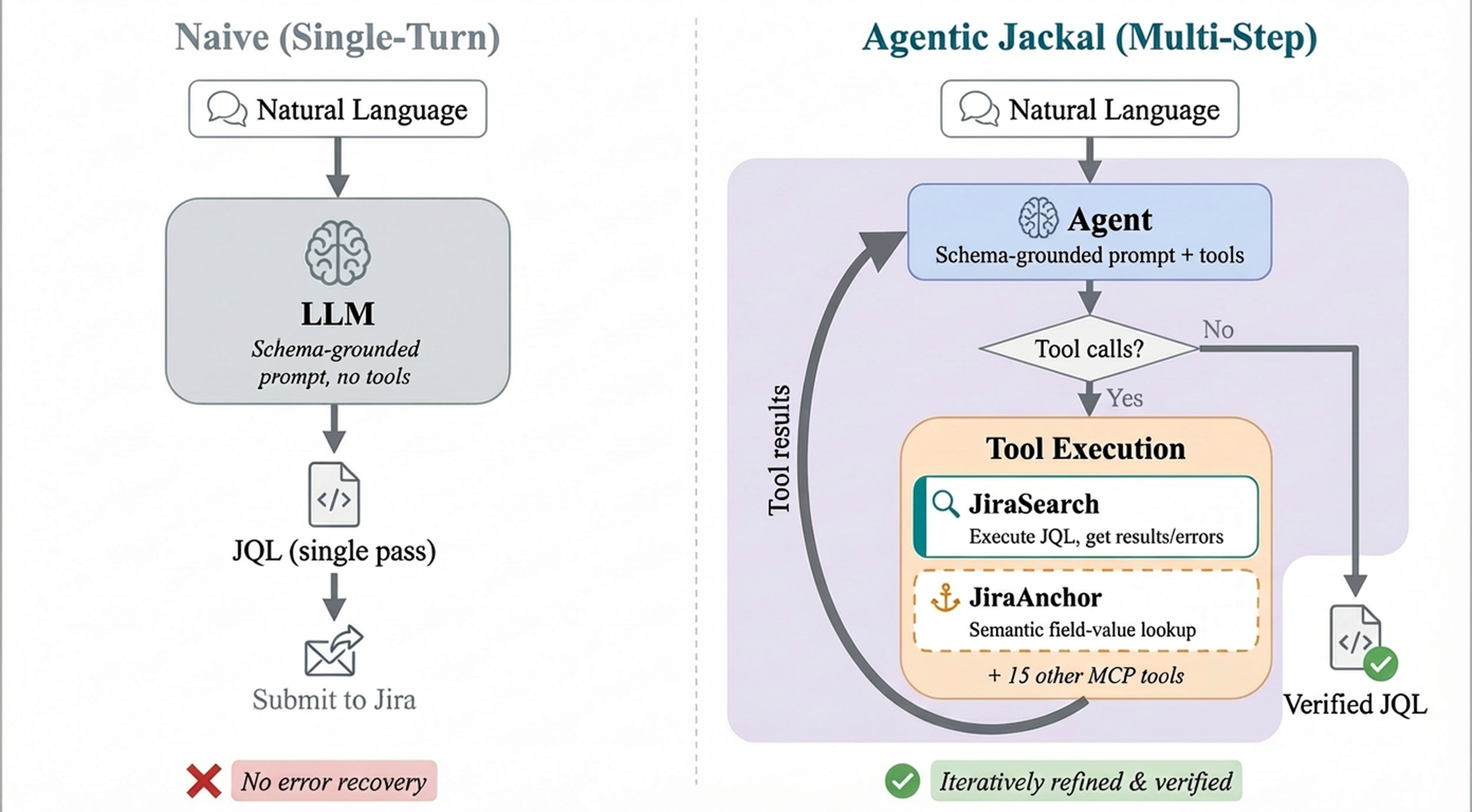}
\caption{Naive single-turn approach (left) compared to Agentic Jackal (right). The naive approach generates JQL in a single pass without tool access. The agent iteratively refines queries using JiraSearch and JiraAnchor (both available in Experiments~1 and~2), looping until it produces a verified output. The JiraAnchor tool enables the agent to resolve ambiguous user queries by retrieving relevant unique values in a field using the live Jira instance.}
\label{fig:naive_compared_agentic}
\end{figure*}

Despite this progress, the Jackal
benchmark~\citep{frank2025jackal}, the first open, execution-based
benchmark for text-to-JQL, reveals that single-turn LLMs achieve
over 90\% accuracy on literal translations but collapse below 30\%
on paraphrased or under-specified variants. These results expose two
structural gaps that single-turn prompting alone cannot close. First,
without a feedback loop, models generating JQL in a single pass
cannot verify or refine their output against the live instance.
Second, categorical fields such as components, fix versions, and labels can contain thousands of instance-specific values, and without a mechanism to discover which values actually exist in the live instance, a user request for "version 6.5" may match none of the actual values (e.g., \texttt{6.5.0}, \texttt{6.5}), causing the generated filter to silently return empty results. Concurrent findings in text-to-SQL confirm
the same two gaps: Spider~2.0~\citep{lei2025spider} shows that
enterprise conditions require iterative execution and refinement, and
DIVER~\citep{nan2026diver} demonstrates that resolving categorical
values without live database access causes significant accuracy drops.

In this paper, we introduce Agentic Jackal, a novel
tool-augmented multi-step JQL agent built on the Jira MCP server that
directly addresses both gaps
(Figure~\ref{fig:naive_compared_agentic}). The agent generates, executes,
and iteratively refines JQL queries using live Jira feedback, where
three distinct signals (a non-empty result set, a zero-result
response, and an error message) enable targeted corrections rather
than speculative rewrites. To resolve the field-value disambiguation
problem, we introduce JiraAnchor, a novel semantic
retrieval tool that queries the live Jira instance using
grep-based and embedding-based similarity search to ground natural language
field-value mentions in actual instance values. Together, these two
components separate the concerns of value grounding and query
verification, each targeting one of the two failure modes from prior approaches.

We evaluate 9 frontier LLMs on Jackal-1K, a 1,000-query stratified
subset of the Jackal benchmark, across two experiments measuring
execution accuracy. The full agentic approach improves 7 of 9 LLMs, with gains concentrated on the most linguistically challenging
query variants (9.0\% relative improvement on Short NL). A controlled ablation
isolating JiraAnchor on field-value-sensitive queries shows accuracy
improving from 48.7\% to 71.7\%, with component accuracy rising from
16.9\% to 66.2\% across all 9 models. A manual error analysis reveals that semantic interpretation ambiguities (issue type disambiguation, text field selection, and version confusion) account for 58\% and 68\% of failures on the two hardest query variants, indicating that dominant errors stem from inherent natural language ambiguity rather than value resolution failures.

\noindent Our contributions are as follows:
\begin{itemize}
    \item \textbf{Agentic Jackal}: A tool-augmented multi-step agent
    built on the Jira MCP server that equips LLMs with live JQL
    execution, iterative query refinement, and semantic value
    grounding via JiraAnchor, establishing the first open agentic
    baseline for enterprise text-to-JQL. We evaluate 9 frontier LLMs
    across two experiments on Jackal-1K, releasing all agent
    transcripts, evaluation results, and framework code.

    \item \textbf{JiraAnchor}: A novel semantic field-value retrieval
    tool that resolves natural language mentions of instance-specific
    categorical values against a live Jira instance, directly
    addressing the user query value disambiguation failure mode in text-to-JQL.

    \item \textbf{Error taxonomy}: We identify
    six dominant failure categories across models and query variants,
    revealing that semantic interpretation ambiguities (issue type
    disambiguation, text field selection, version confusion) account
    for the majority of errors on the hardest variants, rather than
    value resolution failures.
\end{itemize}

% ===========================================================================
% SECTION 2: RELATED WORKS
% ===========================================================================
\section{Related Work}

\subsection{Text-to-SQL and the Enterprise Gap}

Increasingly realistic benchmarks drive progress in natural language
interfaces to databases. WikiSQL~\citep{zhong2017seq2sql} introduced
large-scale, execution-based evaluation on single tables, while
Spider~\citep{yu-etal-2018-spider} shifted the focus to cross-domain
generalization across unseen schemas. BIRD~\citep{li-etal-2023-bird}
scaled to larger, more complex databases with real-world content,
reporting execution accuracy around 73\% for leading systems. Most
recently, Spider~2.0~\citep{lei2025spider} reframes text-to-SQL
around enterprise workflows with large schemas, multiple SQL dialects,
and multi-step interactions. Execution accuracy drops to roughly 21\%,
compared to over 90\% on Spider's single-turn academic setup. This
gap, which the authors attribute to the need for iterative execution,
error recovery, and tool-assisted exploration, parallels the pattern
in Jackal's text-to-JQL evaluation, where single-turn LLMs achieve
over 90\% on literal translations but below 30\% on paraphrased
variants.

\subsection{Agentic Query Generation and Self-Correction}

This enterprise gap motivates multi-step architectures that decompose
query generation and iteratively refine outputs.
DIN-SQL~\citep{pourreza2023dinsql} decomposes text-to-SQL into
sub-tasks including schema linking, query classification, generation,
and self-correction, improving execution accuracy on Spider by roughly
10 percentage points over few-shot prompting. However, DIN-SQL's
self-correction is intrinsic: the model re-reads its own SQL without
executing it. Ablations show this step contributes only 1--3
percentage points, illustrating the limits of model-internal review
without external feedback.

Subsequent systems introduce execution-based feedback loops.
MAC-SQL~\citep{wang2024macsql} employs a multi-agent framework whose
Refiner executes candidate SQL, observes errors or empty result sets,
and rewrites the query accordingly.
CHESS~\citep{talaei2024chess} further specializes for enterprise-scale
databases, pairing an Information Retriever that grounds generation in
actual stored values with a Unit Tester that validates candidates
against generated tests, achieving state-of-the-art accuracy on BIRD.
Both systems confirm that using execution results rather than model
introspection as the refinement signal yields substantially better
accuracy.

These findings align with the broader paradigm of tool-augmented LLM
agents~\citep{lumer2025tool,10.1007/978-3-032-15632-7_2,yao2023react,huang2025_filesystem_context_engineering,lumer2025tooltoagent,patil2025bfcl, gulati2026rowsreasoningagenticretrieval} and 
self-correction~\citep{kamoi2024selfcorrection}, which concludes that
self-correction without external feedback does not reliably improve
outputs, whereas self-correction with execution results or tool
outputs consistently does. Our agent operates in this setting: the
live Jira instance provides unambiguous feedback
signals (non-empty results, zero results, or error
messages), enabling targeted corrections rather than speculative
rewrites.

\subsection{Value Resolution in Structured Query Generation}

Beyond iterative execution, a second structural challenge in
text-to-structured-query systems is resolving natural language mentions
to the exact values stored in the target system. Models trained on
general corpora cannot predict instance-specific values such as
version names, component labels, or project-specific tags. The BIRD
benchmark demonstrates that providing external knowledge hints,
including sampled database values, significantly improves generation
accuracy~\citep{li-etal-2023-bird}, and even frontier LLMs frequently
generate plausible but non-existent values when this evidence is
withheld.

DIVER~\citep{nan2026diver} provides the most thorough treatment of
this problem, showing that accuracy drops by over 10 percentage points
on BIRD when models must resolve categorical values without
expert-provided evidence. DIVER addresses this through a multi-agent
framework for dynamic interactive value linking, where specialized
assistants iteratively probe the database using semantic similarity
search, exact match, and value enumeration to produce verified
evidence for the downstream SQL generator. Additional work shows non-vector strategies in advanced retrieval-augmented generation such as grep-, knowledge-graph-, and table-of-contents-based strategies provide alternatives to vector-only retrieval systems~\cite{lumer2025rethinking,huang2025_filesystem_context_engineering,lumer2025memtool,lumer2024toolshed,heule2025_semantic_search_agents,lumer2025tool,nizar2025agent}.

JiraAnchor targets the same fundamental problem in the JQL domain,
where categorical field values such as components and versions follow
non-obvious, instance-specific naming conventions. Where DIVER employs
multi-turn iterative exploration, JiraAnchor consolidates value
resolution into a single-call retrieval step, prioritizing simplicity
and low latency within the broader agent loop (see
Section~\ref{sec:tools} for implementation details). Both approaches
confirm the same core finding: without live, query-time access to
instance-specific values, model accuracy on value-sensitive queries
collapses. We evaluate JiraAnchor's impact in
Experiment~2 (Section~\ref{sec:exp2}).

\subsection{Text-to-JQL}

Despite the iterative-execution and value-resolution challenges
identified above, open research on text-to-JQL remains limited.
Existing resources include two public community datasets,
Text2JQL~\citep{kulkarni2023text2jql} and
Text2JQL\_v2~\citep{kulkarni2023text2jql_v2}, which are limited in
scale and lack execution-based evaluation, and an enterprise study
reporting 54\% exact match on 218
prompts~\citep{wang2025enterpriselargelanguagemodel}, with neither
data nor scripts released. Proprietary systems such as Atlassian
Intelligence~\citep{atlassian2025intelligence, clovity2025} provide
natural language search but do not publish training data or evaluation
suites.
Jackal~\citep{frank2025jackal} addresses these gaps as the first
open, large-scale, execution-based benchmark for text-to-JQL; its
evaluation exposes two structural failure modes, the inability to
verify output without a feedback loop and unresolved
instance-specific values, that Agentic Jackal directly targets.

% ===========================================================================
% SECTION 3: DATASETS
% ===========================================================================
\section{Datasets}\label{sec:datasets}

\begin{figure*}[t]
\centering
\includegraphics[width=.75\textwidth]{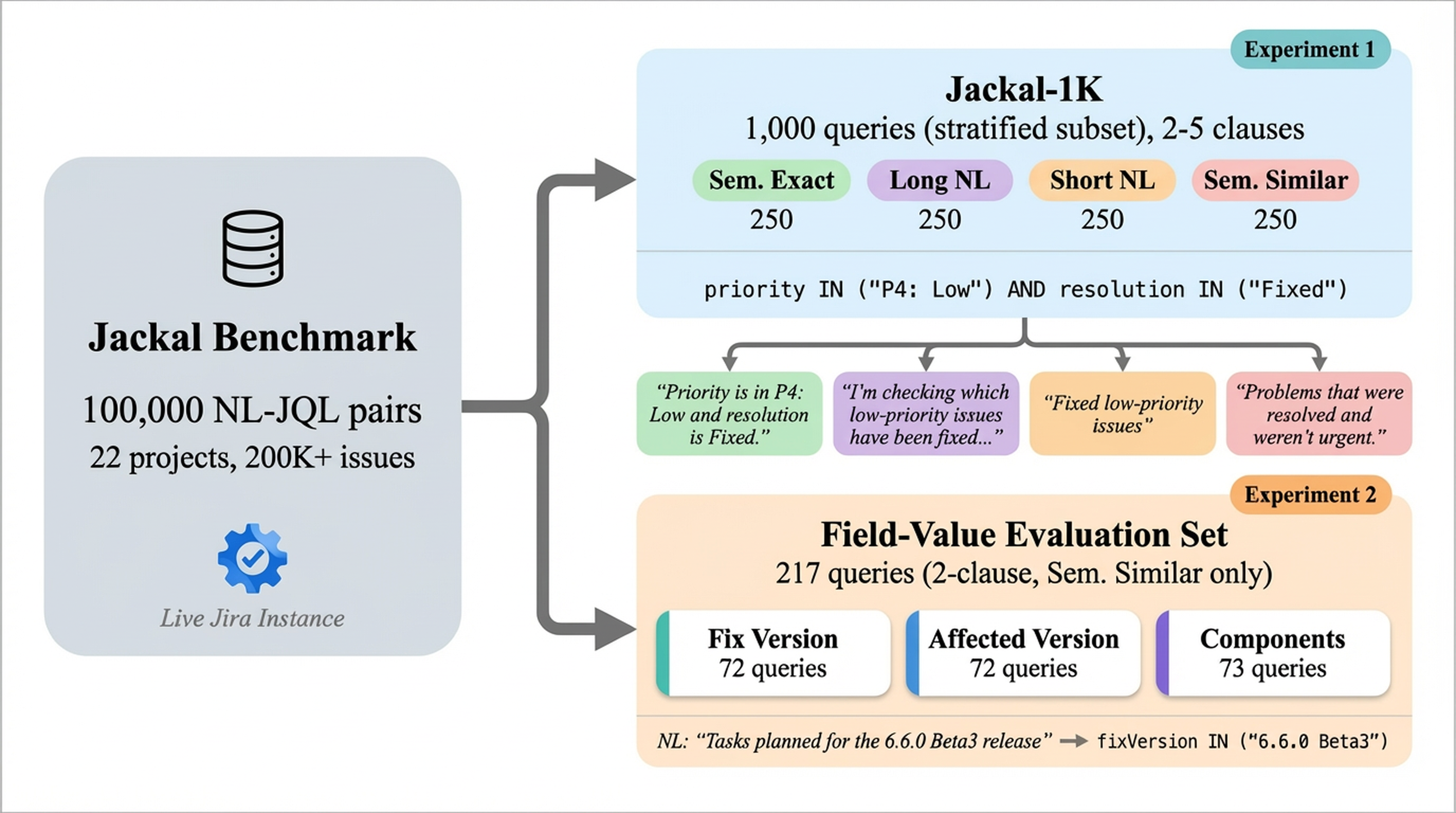}
\caption{Overview of the two evaluation datasets. Left: Jackal-1K, a 1,000-query stratified subset of the Jackal benchmark with four NL variants per query, used in Experiment~1. Right: The Field-Value Evaluation Set, a targeted set of 2-clause queries covering three categorical fields, used in Experiment~2.}
\label{fig:datasets}
\end{figure*}

\subsection{Jackal-1K}\label{sec:jackal1k}

We draw our primary evaluation data from
Jackal~\citep{frank2025jackal}, the first open, execution-based
benchmark for text-to-JQL. The full benchmark pairs 100,000 natural
language requests with validated JQL queries on a live Jira instance
containing over 200,000 issues spanning 22 projects. We select a
stratified 1,000-query subset, Jackal-1K, as our evaluation set.
Agentic evaluation requires live API calls for every query, model, and
experimental condition; at 9 models and two conditions per experiment,
a larger subset such as Jackal-5K would be cost- and rate-limit-prohibitive.
We verify that Jackal-1K preserves the clause-count, field-type, and
variant distributions of Jackal-5K from the original benchmark, ensuring
that results generalize.

Queries are constructed programmatically by assembling multi-clause
Boolean filters (2--5 clauses) from field-operator-value triples
drawn from the live instance schema. Hand-crafted rules filter out
implausible combinations, and every surviving query is executed
against the live instance; any that return an empty result set are
discarded. The remaining queries are sampled with balancing
constraints across categorical value groups, date ranges, and text
fields to ensure broad schema coverage (see~\citet{frank2025jackal}
for the full construction pipeline).

Each validated query is paired with four natural language
formulations produced by an LLM under controlled style
prompts: \textit{Semantically Exact}, a word-for-word rendering of
the JQL logic that provides an upper bound on model
performance; \textit{Long NL}, a multi-sentence explanation that
embeds each clause in a broader context; \textit{Short NL}, a terse
phrase naming the user's intent without referencing JQL fields or
operators; and \textit{Semantically Similar}, a paraphrase that
avoids all JQL vocabulary, requiring the model to recover both the
correct field and the correct value from context alone.
Appendix~\ref{appendix:jql_examples} provides full examples at
varying clause complexities, and
Appendix~\ref{appendix:query_variations} reproduces the prompt
templates used to generate each variant.

From this pool, Jackal-1K draws 250 queries per variant, balanced
across clause counts, yielding 1,000 (NL, JQL) pairs. The mean
clause count is 3.5 (range 1--6), and average request length ranges
from 5.9 words (Short NL) to 36.4 words (Long NL), with
Semantically Exact at 17.8 and Semantically Similar at 16.6. The
queries span 15 distinct Jira fields covering project metadata,
dates, free-text search, and categorical attributes. Jackal-1K is used in Experiment~1
(Figure~\ref{fig:datasets}, left).

\subsection{Field-Value Evaluation Set}\label{sec:fv-eval}

To isolate the impact of JiraAnchor on categorical field-value
resolution, we construct a separate, targeted evaluation set that is
entirely disjoint from Jackal-1K.
Whereas Jackal-1K queries contain 2--5 clauses spanning many fields,
this set uses minimal 2-clause queries that pair a project filter
with exactly one categorical field, ensuring that execution accuracy
depends almost entirely on whether the model resolves the field value
correctly.

Each query follows the pattern
\texttt{project IN ("<project>") AND <field> IN ("<value>")} and is
expressed in the Semantically Similar variant, where the natural
language avoids reusing field names or stored values. We chose
Semantically Similar because it requires the most value inference
and is where JiraAnchor provides the clearest signal.

The set covers three categorical fields:

\begin{itemize}
    \item \textbf{Fix Version} (72 queries): e.g.,
    \textit{``Tasks from the QTBUG initiative planned for the 6.6.0
    Beta3 release''} $\rightarrow$
    \texttt{fixVersion IN ("6.6.0 Beta3")}

    \item \textbf{Affected Version} (72 queries): e.g.,
    \textit{``Problems encountered in the first beta of version
    6.4.0''} $\rightarrow$
    \texttt{affectedVersion IN ("6.4.0 Beta1")}

    \item \textbf{Components} (73 queries): e.g.,
    \textit{``Tasks related to data structures and their
    operations''} $\rightarrow$
    \texttt{component IN ("Core: Containers and Algorithms")}
\end{itemize}

The component example illustrates the challenge: the natural language
description ``data structures and their operations'' must be resolved
to the exact stored name ``Core: Containers and Algorithms,'' a
mapping that cannot be inferred without querying the live instance.
This evaluation set is used in Experiment~2
(Figure~\ref{fig:datasets}, right).

\section{Agentic Jackal Architecture}\label{sec:system}

Agentic Jackal is a tool-augmented multi-step agent that generates,
executes, and iteratively refines JQL queries using live Jira
feedback. Given a natural language request, the agent produces a
candidate JQL query, executes it against the live Jira instance,
observes the result, and decides whether to refine the query or
return a final answer. The system comprises three components: the
agent loop (Section \ref{subsec:agent-loop}), which orchestrates iterative
query refinement; Jira Search (Section \ref{subsec:tools}), which provides
live execution feedback; and JiraAnchor (Section \ref{sec:tools}), a
semantic retrieval tool that resolves natural language value mentions
to their canonical stored forms
(Figure~\ref{fig:naive_compared_agentic}).

\subsection{Agent Loop}\label{subsec:agent-loop}

The agent is implemented as a two-node directed graph. The first
node contains the LLM with a schema-grounded system prompt and bound
tool definitions. The second node executes any tool calls issued by
the LLM and returns results. Routing is conditional: if the LLM's
response contains one or more tool calls, control passes to the tool
execution node and then returns to the LLM; if no tool calls are
present, the graph terminates and the agent's final message is
returned. This pattern allows the agent to chain multiple tool calls
across successive iterations before producing a final answer.
The loop terminates when the LLM issues a response containing no
tool calls, which serves as an implicit ``accept'' signal: the model
has judged the most recent execution result sufficient and returns
the final JQL. A framework-level recursion limit of 25 graph steps
(approximately 12 tool-calling iterations) acts as a hard ceiling;
if the agent exhausts this budget without converging, the run is
recorded as a failure and no query is returned.

The agent is equipped with 17 tools: 16 from the Atlassian Jira MCP
server~\citep{atlassian2025mcp} and one custom tool
(JiraAnchor, Subsection \ref{sec:tools}). For text-to-JQL, the agent
primarily uses Jira Search and JiraAnchor; the remaining MCP
operations are bound but unused during evaluation. The agent is
model-agnostic: all evaluated LLMs share the same architecture,
system prompt, and tool definitions, isolating model capability from
system design.

\begin{figure}[t]
   \centering
   \includegraphics[width=0.66\columnwidth]{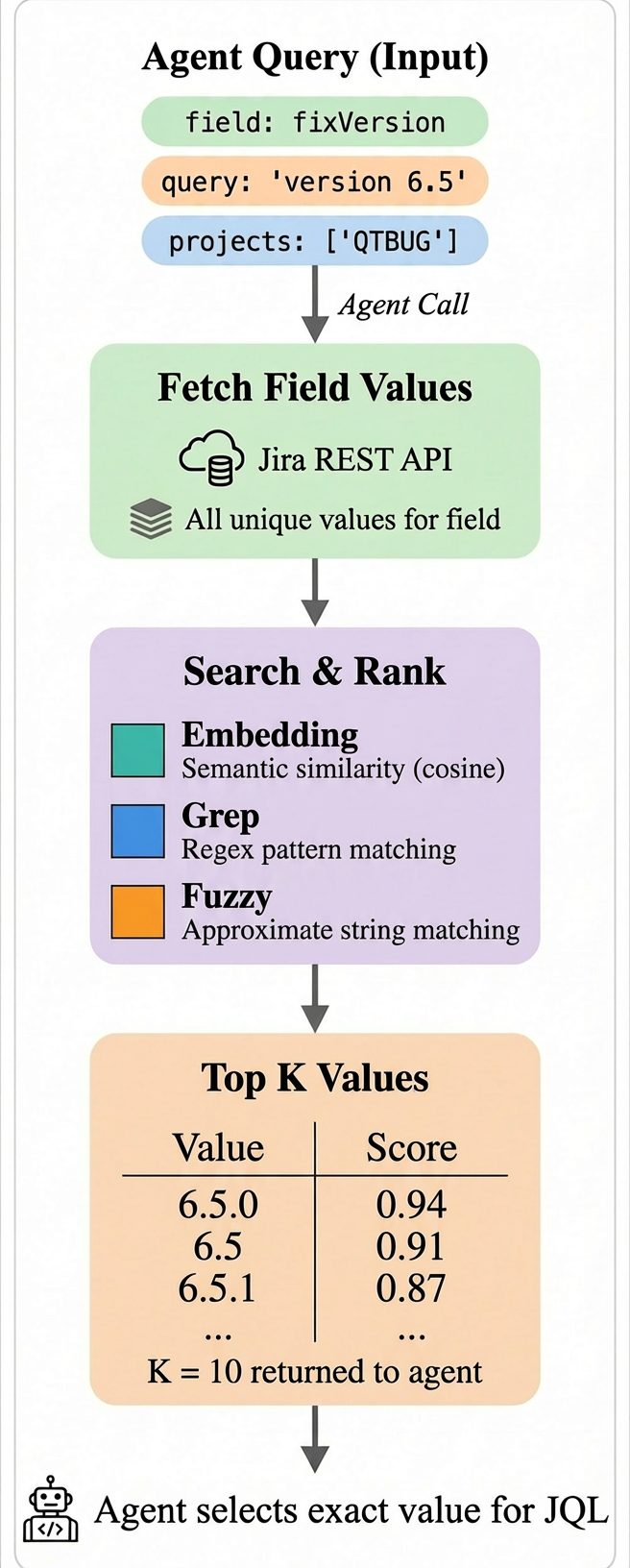}
   \caption{JiraAnchor pipeline. Given a field name and natural language value mention, the tool fetches candidate values from the live Jira instance, computes embedding-based similarity scores, and returns top-10 matches ($K{=}10$) ranked by cosine similarity.}
   \label{fig:jiraanchor}
\end{figure}

\subsection{Jira Search}\label{subsec:tools}

The primary execution tool accepts a JQL query string and runs it
against the live Jira instance via the MCP server. The response
includes matching issue keys, the total result count, and, for
invalid queries, the Jira error message describing the syntax or
schema violation. This feedback is returned directly to the LLM,
giving the agent three distinct signals: a non-empty result set
confirms the query is plausible, a zero-result response suggests an
overly restrictive or misaligned filter, and an error message
pinpoints the exact clause that failed. The agent uses these signals
to decide whether to return the current query, relax constraints, or
correct a field name or value before re-executing.

\subsection{JiraAnchor}\label{sec:tools}

While Jira Search validates query correctness through execution
feedback, it cannot resolve natural language mentions of
instance-specific categorical values. JiraAnchor is a custom
semantic field-value retrieval tool that we implement outside
the Atlassian MCP server to address this value disambiguation
problem. It connects to the same Jira instance as the MCP tools
but is not part of the MCP server's tool suite; it is a standalone
component purpose-built for text-to-JQL
(Figure~\ref{fig:jiraanchor}).

JiraAnchor accepts three inputs: a JQL field name (e.g.,
\texttt{fixVersion}, \texttt{component}, \texttt{labels}), a natural
language query describing the value the user intends, and an optional
list of project keys to scope the search. It operates in two stages
(Figure~\ref{fig:jiraanchor}). First, it fetches all unique values
for the specified field from the live Jira instance via the REST API,
using paginated requests scoped to the provided projects. Second, it
ranks all candidate values against the query using one of several
configurable search strategies, including semantic embedding, regex
pattern matching, and approximate string matching, and returns the
top-$K$ matches ($K{=}10$) with their exact names, ready for direct
use in JQL. In our experiments we use the embedding strategy, which
computes cosine similarity between the query and candidate value
embeddings.

For example, given the query ``Tasks planned for version 6.5,'' the
agent calls
\texttt{JiraAnchor(field="fixVersion", query="6.5",
projects=["QTBUG"])} and receives matches such as \texttt{"6.5"},
\texttt{"6.5.0"}, \texttt{"6.5.1"}, and \texttt{"6.5.0 Beta1"},
allowing it to select the correct value rather than guessing.

\subsection{Prompt Design}

Beyond the tools, the system prompt plays a central role in guiding
the agent's query construction decisions. The prompt grounds the
agent in the target Jira instance by
providing a structured JSON schema containing all JQL field names,
supported operators, and enumerated values for fields with fixed
vocabularies. For example, \texttt{issuetype} lists all 15 valid
issue types, \texttt{priority} lists all 7 priority levels, and
\texttt{resolution} lists all 9 resolution values
(Appendix~\ref{appendix:jql_schema} lists all 15 fields and their
attributes). This eliminates guesswork for fields with small, known
value sets.

For fields with large or instance-specific value sets, the schema
explicitly marks values as truncated and instructs the agent to use
JiraAnchor rather than synthesizing values. The fields
\texttt{component}, \texttt{fixVersion}, \texttt{affectedVersion},
and \texttt{labels} all carry this annotation. This design separates
the concerns of value grounding (handled by JiraAnchor) from query
construction (handled by the LLM), ensuring that the agent consults
the live instance for values it cannot know from static context.

The prompt also provides field aliases that normalize common natural
language mentions to their canonical JQL names, such as ``fix
version'' to \texttt{fixVersion} and ``components'' to
\texttt{component}. The agent is instructed to always execute
queries via Jira Search rather than returning JQL as plain text, and
to make reasonable assumptions rather than asking clarifying
questions. The full system prompt is provided in
Appendix~\ref{appendix:system_prompt}.

% ===========================================================================
% SECTION 5: EXPERIMENTS
% ===========================================================================

% Old Table 1 removed -- restructured into simplified Table 1 inside Sec 5.2

\begin{figure*}[t]
\centering
\includegraphics[width=1.8\columnwidth]{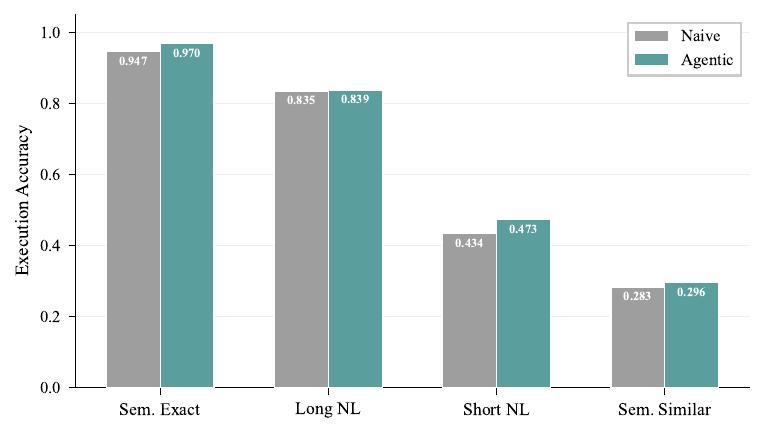}
\caption{Average execution accuracy by query variant (9 models). Short NL shows the largest agentic gain (improvement of 3.9\%), while Sem.\ Exact is near ceiling for both conditions (improvement of 2.3\%).}
\label{fig:exp1_variants}
\end{figure*}

\section{Experiments}\label{sec:experiments}

\subsection{Experimental Setup}\label{sec:setup}

\paragraph{Datasets.}
Two evaluation sets are used, both derived from the Jackal
benchmark~\citep{frank2025jackal}. Jackal-1K
(Section~\ref{sec:jackal1k}) is a stratified 1,000-query subset with
250 queries per NL variant (Long NL, Short NL, Semantically Exact,
Semantically Similar), balanced across clause counts; it is used in
Experiment~1 to evaluate overall agentic execution accuracy. The
Field-Value Evaluation Set (Section~\ref{sec:fv-eval}) contains 217
minimal 2-clause queries spanning three categorical fields
(\texttt{fixVersion}, \texttt{affectedVersion}, \texttt{component}),
entirely disjoint from Jackal-1K; it is used in Experiment~2 to
isolate the contribution of JiraAnchor.

\paragraph{Baselines.}
Each experiment compares the proposed system against a controlled
ablation. In Experiment~1, the baseline is \textit{naive single-turn
generation}: the model receives a schema-grounded prompt and produces
JQL in a single forward pass without tool access or execution
feedback, following the original Jackal evaluation
protocol~\citep{frank2025jackal}. In Experiment~2, the baseline is
the full agentic system with JiraAnchor removed: all prompt references
to JiraAnchor are stripped and the tool is unbound, while all other
MCP tools remain available.

\paragraph{Metrics.}
All experiments use \textbf{execution accuracy (EX)} as the primary
metric. A predicted JQL query is correct if, when executed against
the live Jira instance, it returns the same set of issue keys as the
gold query. This metric is robust to syntactic variation and
semantically equivalent rewrites, directly measuring whether the
generated query retrieves the intended issues.

\paragraph{Models.}
We evaluate 9 frontier LLMs spanning four providers in both experiments: GPT-5.2,
GPT-4o, GPT-5 Mini, and GPT-OSS 120B (OpenAI); Claude Sonnet 4.5
and Claude Opus 4.6 (Anthropic); Gemini 3 Flash and Gemini 3 Pro
(Google); and Pixtral Large (Mistral).

\paragraph{Infrastructure.}
All agentic runs use the architecture described in
Section~\ref{sec:system}, executed within the same agent 
environment. Inference uses each provider's default decoding
settings with temperature 0 where supported. Each query is evaluated
in a single run (no majority voting or repeated sampling). The two
experiments use different prompts tailored to their respective
evaluation goals (see below).

\subsection{Experiment 1: Naive Compared to Agentic}\label{sec:exp1}

Each model is evaluated on Jackal-1K
(Section~\ref{sec:jackal1k}) under two conditions with
different prompts. In the \textit{naive} condition, the
model receives the original Jackal benchmark
prompt~\citep{frank2025jackal}: a schema-grounded system prompt
containing JQL field definitions, operators, and enumerated values,
which instructs the model to generate JQL directly as plain text in
a single pass without tool access. In the \textit{agentic}
condition, the model receives a tool-augmented prompt
(Section~\ref{sec:system}) that shares the same underlying JQL
schema but additionally includes tool-use instructions (directing
the model to execute queries via Jira Search and to call JiraAnchor
for categorical value resolution), field aliases, and chain-of-thought
scaffolding for ambiguous field selection. The naive condition was
re-evaluated by us on the Jackal-1K subset using the original prompt
to ensure an apples-to-apples comparison on the same queries.
Table~\ref{tab:exp1} reports results.

\begin{table}[t]
\centering
\small
\setlength{\tabcolsep}{5pt}
\begin{tabular}{l|cc|c}
\toprule
\textbf{Model} & \textbf{Naive} & \textbf{Agentic} & \textbf{$\Delta$} \\
\midrule
Gemini 3 Flash      & .628 & \textbf{.710} & \textbf{+.082} \\
Gemini 3 Pro        & .641 & .671          & +.030 \\
GPT-5 Mini          & .616 & .638          & +.022 \\
Claude Opus 4.6     & .634 & .653          & +.019 \\
Pixtral Large       & .537 & .554          & +.017 \\
GPT-5.2             & \textbf{.659} & .670 & +.011 \\
Claude Sonnet 4.5   & .633 & .644          & +.011 \\
GPT-OSS 120B        & .630 & .624          & $-$.006 \\
GPT-4o              & .643 & .634          & $-$.009 \\
\midrule
\textit{Average}    & \textit{.625} & \textit{.644} & \textit{+.020} \\
\bottomrule
\end{tabular}
\caption{Overall execution accuracy, naive compared to agentic on Jackal-1K. Seven of nine models improve. Table~\ref{tab:exp1_full} (Appendix) provides full per-variant detail.}
\label{tab:exp1}
\end{table}

\begin{figure*}[t]
\centering
\includegraphics[width=1.4\columnwidth]{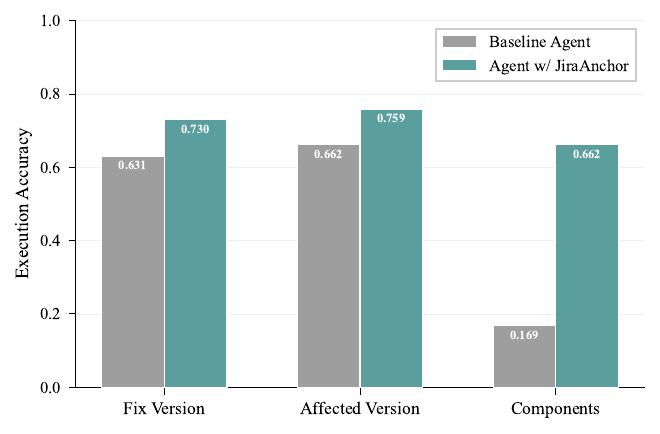}
\caption{Average execution accuracy by categorical field (9 models). JiraAnchor yields the largest gain on component queries (improvement of 49.3\%), where baseline accuracy without value grounding collapses to 16.9\%.}
\label{fig:exp2_fields}
\end{figure*}

Across models, the agentic approach improves the average from 62.5\%
to 64.4\% overall, an improvement of 2.0\%, with gains concentrated on Short NL
(improvement of 3.9\%). The best individual result is Gemini 3 Flash, which
improves from 62.8\% to 71.0\% overall, an improvement of 8.2\%, driven by a large
gain on Short NL (improvement of 17.3\%). Seven of nine models improve under the
agentic approach, while two show slight regressions; we analyze
per-variant patterns and failure cases in
Section~\ref{sec:error_taxonomy}.

\subsection{Experiment 2: JiraAnchor Ablation}\label{sec:exp2}

To isolate JiraAnchor's contribution to field-value resolution, we
run a controlled ablation on the field-value evaluation set
(Section~\ref{sec:fv-eval}). Each model is evaluated under two
conditions. Both conditions use a minimal, field-specific prompt
scoped to the target categorical field (e.g., \texttt{project} +
\texttt{fixVersion} only). In the \textit{with JiraAnchor}
condition, the prompt instructs the model to call JiraAnchor to
resolve the field value before constructing JQL, and the tool is
bound to the agent. In the \textit{without JiraAnchor} condition,
all JiraAnchor references are removed from the prompt text (e.g.,
\texttt{valuesNote} changes from ``Use search\_jira\_values to find
exact values'' to ``use exact value from user query''), and the tool
is removed from the agent's tool bindings. All other MCP tools
remain available in both conditions. This design cleanly isolates
the marginal contribution of semantic field-value retrieval from the
base agentic capability.
Table~\ref{tab:exp2} reports results across 9 models and three
categorical fields.

\begin{table}[t]
\centering
\small
\setlength{\tabcolsep}{5pt}
\begin{tabular}{l|cc|c}
\toprule
\textbf{Model} & \textbf{Baseline} & \textbf{w/ JiraAnchor} & \textbf{$\Delta$} \\
\midrule
GPT-5 Mini           & .393 & .710 & \textbf{+.317} \\
GPT-5.2              & .416 & .705 & +.289           \\
Pixtral Large        & .453 & \textbf{.742} & +.289  \\
Claude Sonnet 4.5    & .462 & .733 & +.271           \\
GPT-4o               & .449 & .710 & +.261           \\
Gemini 3 Pro         & .527 & \textbf{.742} & +.215  \\
GPT-OSS 120B         & .458 & .645 & +.187           \\
Gemini 3 Flash       & .601 & .733 & +.132           \\
Claude Opus 4.6      & \textbf{.628} & .733 & +.105  \\
\midrule
\textit{Average}     & \textit{.487} & \textit{.717} & \textit{+.230} \\
\bottomrule
\end{tabular}
\caption{Overall execution accuracy with and without JiraAnchor across 9 models. All models show positive gains. Table~\ref{tab:exp2_full} (Appendix) provides full per-field detail.}
\label{tab:exp2}
\end{table}

Overall, JiraAnchor improves accuracy from 48.7\% to 71.7\%. The largest gain occurs on
\texttt{component} queries, where accuracy rises from 16.9\% to
66.2\%. Version fields show moderate but consistent
improvements: \texttt{affectedVersion} from 66.2\% to 75.9\%
and \texttt{fixVersion} from 63.1\% to 73.0\%.
All 9 models show positive overall deltas, with GPT-5 Mini gaining
the most (improvement of 31.7\%).

\begin{table}
\centering
\scriptsize
\setlength{\tabcolsep}{3pt}
\begin{tabular}{@{}p{2.2cm}p{2.5cm}p{2.5cm}@{}}
\toprule
\textbf{NL Mention} & \textbf{Without} & \textbf{With JiraAnchor} \\
\midrule
``7.0'' \newline {\scriptsize field: fixVersion} &
\texttt{``7.0''} \newline {\scriptsize 0 results} &
\texttt{``7.0 (Next Major Release)''} \newline {\scriptsize correct} \\
\addlinespace[4pt]
``build tools'' \newline {\scriptsize field: component} &
\texttt{``build tools''} \newline {\scriptsize 0 results} &
\texttt{``Build tools: Other''} \newline {\scriptsize correct} \\
\addlinespace[4pt]
``QDS 1.6 Beta'' \newline {\scriptsize field: fixVersion} &
{\scriptsize not resolved} &
\texttt{``QDS 1.6 Beta''} \newline {\scriptsize correct} \\
\bottomrule
\end{tabular}
\caption{JiraAnchor success cases. Each row isolates a single field-value mention. Without JiraAnchor, the agent guesses a value that does not exist; with it, the canonical stored form is retrieved.}
\label{tab:examples}
\end{table}

\begin{figure*}[!b]
\centering
\includegraphics[width=\textwidth]{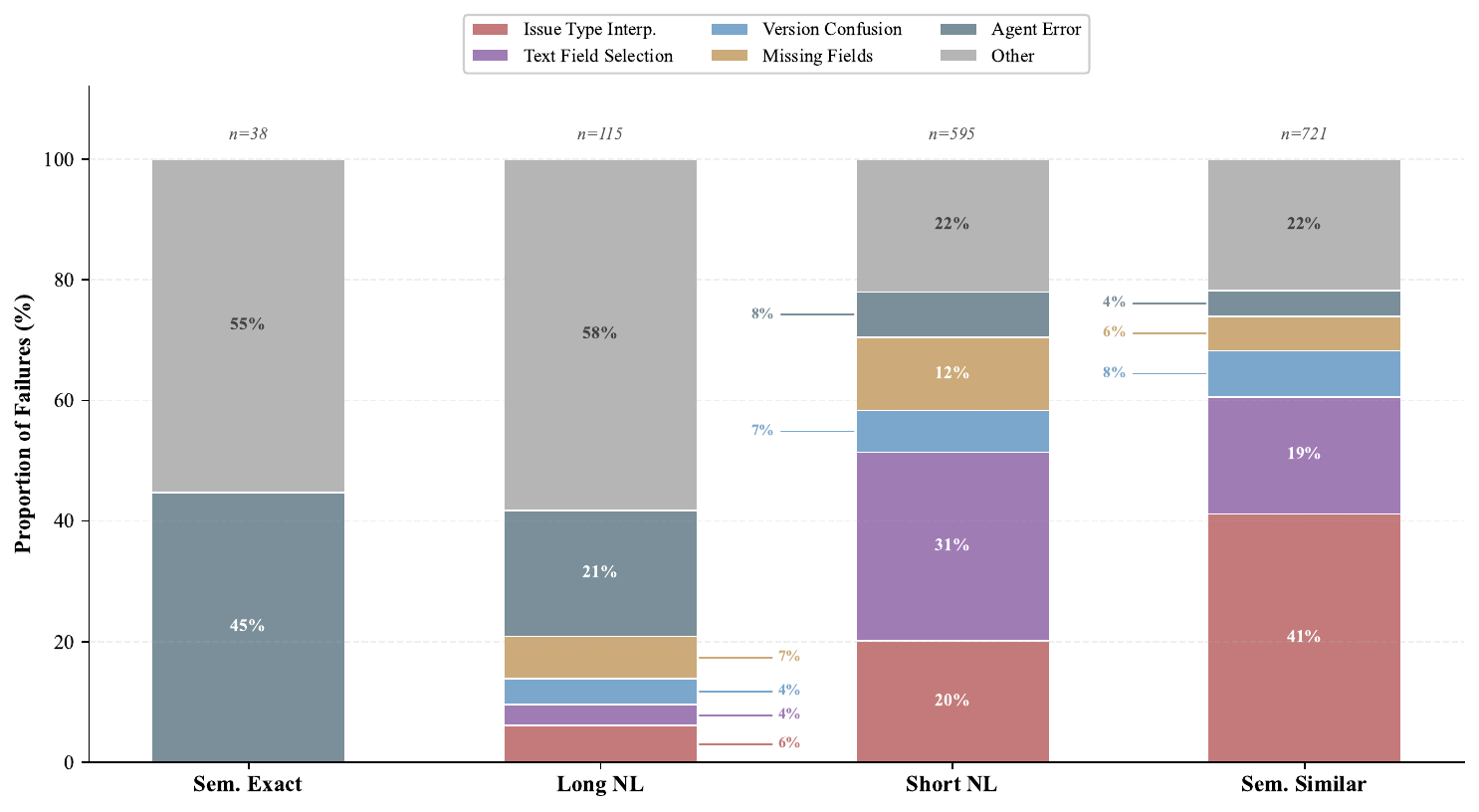}
\caption{Error taxonomy for Agentic Jackal failures by query variant. Each bar shows the proportion of failures attributed to six error categories. On the hardest variants (Short NL and Sem.\ Similar), semantic interpretation errors dominate: \textbf{Issue Type Interp.} (e.g., ``bugs'' as \texttt{issuetype=Bug} vs.\ text search), \textbf{Text Field Selection} (\texttt{summary} vs.\ \texttt{description}), and \textbf{Version Confusion} (\texttt{affectedVersion} vs.\ \texttt{fixVersion}). These are inherent ambiguities that tools cannot resolve.}
\label{fig:error_taxonomy}
\end{figure*}

\paragraph{Impact by field type.}
JiraAnchor is most effective on component queries, where accuracy
without the tool collapses to 16.9\% across models. Component
names in the target Jira instance follow non-obvious naming
conventions (e.g., ``Build tools: Other'', ``QML: Tooling''),
and without access to the live value set, models consistently
generate plausible but incorrect values. JiraAnchor resolves
this by querying the instance for exact component names, enabling
all 9 evaluated models to exceed 61\% accuracy on these queries.

For version fields, the improvement is moderate but consistent.
Version names frequently include suffixes or qualifiers that the
model cannot predict from the natural language alone (e.g., ``7.0
(Next Major Release)'' instead of ``7.0'', ``QDS 1.6 Beta'' instead
of ``1.6''). JiraAnchor retrieves the full value set and matches the
user's mention to the canonical name, recovering queries that would
otherwise return empty result sets.

\paragraph{Per-field regressions.}
While JiraAnchor improves overall accuracy for all 9 evaluated
models, it introduces regressions in specific model-field
combinations. The most notable case is Gemini 3 Flash on fix
versions (71.0\% with JiraAnchor compared to 79.0\% without, a regression of 8.0\%),
even as the same model gains an improvement of 41.3\% on components. We trace
per-field regressions to three patterns.

Overmatching occurs when the tool returns a broader
version name than the gold query expects (e.g., ``4.7'' instead of
``4.7.0''), causing the agent to use the shorter value and return
zero results. Field confusion arises when the tool
correctly identifies a value but the agent assigns it to the wrong
field (e.g., \texttt{affectedVersion} instead of
\texttt{fixVersion}), an ambiguity in the natural language that the
tool exacerbates by increasing the agent's confidence in an
incorrect choice. Unnecessary invocation occurs when the
natural language already contains the exact JQL value, and invoking
JiraAnchor returns a different match, degrading a query that would
have been correct without intervention.

These patterns suggest that selective invocation, where the agent
decides whether to call JiraAnchor based on the confidence of its
initial value, could reduce per-field regressions while preserving
the strong overall gains.

\begin{table*}[t]
\centering
\begin{minipage}[t]{0.48\textwidth}
\centering
\small
\setlength{\tabcolsep}{3pt}
\begin{tabular}{@{}l|cc|cc|c@{}}
\toprule
 & \multicolumn{2}{c|}{\textbf{Naive}} & \multicolumn{2}{c|}{\textbf{Agentic}} & \\
\cmidrule(lr){2-3} \cmidrule(lr){4-5}
\textbf{Model} & \makecell{\textbf{Lat.}\\\textbf{(s)}} & \makecell{\textbf{Tok.}\\\textbf{(avg)}} & \makecell{\textbf{Lat.}\\\textbf{(s)}} & \makecell{\textbf{Tok.}\\\textbf{(avg)}} & \makecell{\textbf{Tool}\\\textbf{Calls}} \\
\midrule
GPT-OSS 120B        & \textbf{2.3}  & 1,928           & 37.6           & 39,340           & 1.7 \\
GPT-4o              & 2.6           & 1,573           & 30.5           & 20,395           & 1.2 \\
Gemini 3 Flash      & 2.8           & 1,532           & 29.9           & 20,085           & 1.3 \\
GPT-5.2             & 2.9           & 1,574           & 29.4           & 17,549           & \textbf{1.0} \\
Pixtral Large       & 3.2           & \textbf{1,487}  & \textbf{20.1}  & \textbf{17,473}  & \textbf{1.0} \\
Claude Sonnet 4.5   & 3.2           & 2,404           & 33.9           & 37,603           & 1.2 \\
Claude Opus 4.6     & 3.7           & 2,395           & 41.3           & 34,582           & 1.5 \\
GPT-5 Mini          & 9.1           & 1,962           & 39.1           & 37,470           & 1.3 \\
Gemini 3 Pro        & 13.5          & 1,746           & 30.2           & 20,140           & 1.2 \\
\midrule
\textit{Average}    & \textit{4.8}  & \textit{1,845}  & \textit{32.4}  & \textit{27,182}  & \textit{1.3} \\
\bottomrule
\end{tabular}
\captionof{table}{Operational cost per query, Experiment~1 (Jackal-1K). Lat.\ = latency, Tok.\ = total tokens. Bold = lowest per column.}
\label{tab:cost-exp1}
\end{minipage}%
\hfill
\begin{minipage}[t]{0.50\textwidth}
\centering
\footnotesize
\setlength{\tabcolsep}{2pt}
\begin{tabular}{@{}l|ccc|ccc@{}}
\toprule
 & \multicolumn{3}{c|}{\textbf{JiraAnchor}} & \multicolumn{3}{c}{\textbf{Baseline}} \\
\cmidrule(lr){2-4} \cmidrule(l){5-7}
\textbf{Model} & \makecell{\textbf{Lat.}\\\textbf{(s)}} & \makecell{\textbf{Tok.}\\\textbf{(avg)}} & \makecell{\textbf{TC}} & \makecell{\textbf{Lat.}\\\textbf{(s)}} & \makecell{\textbf{Tok.}\\\textbf{(avg)}} & \makecell{\textbf{TC}} \\
\midrule
GPT-OSS 120B        & \textbf{26.6}  & 60,964           & 2.0          & 8.6            & 25,318           & 1.1 \\
GPT-5.2             & 36.3           & \textbf{18,546}  & 2.0          & \textbf{8.4}   & \textbf{13,400}  & \textbf{1.0} \\
GPT-4o              & 44.8           & 23,941           & 2.0          & 8.5            & 23,885           & \textbf{1.0} \\
Gemini 3 Flash      & 45.9           & 23,349           & 2.0          & 11.6           & 25,683           & 1.7 \\
Gemini 3 Pro        & 45.1           & 25,077           & 2.1          & 13.8           & 14,578           & 1.0 \\
Claude Sonnet 4.5   & 48.7           & 55,547           & 2.0          & 19.1           & 66,288           & 1.2 \\
Pixtral Large       & 54.0           & 35,721           & 2.0          & 18.9           & 22,176           & 1.0 \\
Claude Opus 4.6     & 69.5           & 69,832           & 2.1          & 38.2           & 65,180           & 1.7 \\
GPT-5 Mini          & 103.6          & 48,560           & 2.0          & 37.9           & 35,319           & 1.0 \\
\midrule
\textit{Average}    & \textit{52.7}  & \textit{40,171}  & \textit{2.0} & \textit{18.3}  & \textit{32,425}  & \textit{1.2} \\
\bottomrule
\end{tabular}
\captionof{table}{Operational cost per query, Experiment~2 (field-value evaluation). Lat.\ = latency, Tok.\ = total tokens, TC = tool calls. Bold = lowest per column.}
\label{tab:cost-exp2}
\end{minipage}
\end{table*}

\subsection{Error Taxonomy}\label{sec:error_taxonomy}

Drawing from a manual analysis of failure cases across models and
query variants, we identify six dominant error categories
(Figure~\ref{fig:error_taxonomy}). All percentages below refer to
the proportion of failures within each query variant.

\begin{itemize}
    \item \textbf{Issue Type Interpretation.} When the user says
    ``bugs,'' the gold JQL uses \texttt{summary \char`~ "bug"} (text
    search), while the agent interprets this as
    \texttt{issuetype = Bug}. This is the single largest category on
    Sem.\ Similar (41\%) and accounts for 20\% of Short NL failures,
    reflecting a dataset convention rather than a genuine agent error.

    \item \textbf{Text Field Selection.} The natural language
    provides no reliable signal to distinguish \texttt{summary} from
    \texttt{description} as the target field for text search. The
    gold JQL assigns these arbitrarily. This category accounts for
    31\% of Short NL and 19\% of Sem.\ Similar failures.

    \item \textbf{Version Confusion.} The agent confuses
    \texttt{affectedVersion} with \texttt{fixVersion} when the
    natural language does not disambiguate (e.g., ``version 6.5'').
    This accounts for 8\% of Sem.\ Similar, 7\% of Short NL, and
    4\% of Long NL failures.

    \item \textbf{Missing Fields.} The agent omits a clause present
    in the gold JQL, typically a secondary filter that the natural
    language implies only weakly. This accounts for 12\% of Short NL,
    7\% of Long NL, and 6\% of Sem.\ Similar failures.

    \item \textbf{Agent Error.} The agent produces a malformed query,
    hits the recursion limit, or fails to call the appropriate tool.
    This accounts for 45\% of Sem.\ Exact failures (second only to
    Other), 21\% of Long NL, and 8\% of Short NL, but is negligible
    on Sem.\ Similar (4\%).

    \item \textbf{Other.} Residual errors including resolution
    mismatches (e.g., \texttt{resolution IS EMPTY} compared to a
    specific set of values), priority mapping ambiguities, and
    semantic paraphrase failures where the agent interprets a synonym
    literally. This category accounts for 22\% of Short NL and
    Sem.\ Similar, 58\% of Long NL, and 55\% of Sem.\ Exact
    failures.
\end{itemize}

A key finding is that the three semantic interpretation
categories (Issue Type Interpretation, Text Field Selection, and
Version Confusion) together account for 58\% of Short NL and 68\%
of Sem.\ Similar failures. These stem from inherent ambiguities in
the natural language or conventions in the benchmark construction,
rather than deficiencies in the agent or its tools. The agent's
tools address value lookup effectively, but the dominant failure
modes are semantic interpretation challenges that require additional
context beyond what any tool can provide.

\subsection{Operational Cost}\label{sec:cost}

Tables~\ref{tab:cost-exp1} and~\ref{tab:cost-exp2} report the
average latency, token usage, and tool call count per query for
each model on Experiments~1 and~2, respectively.

In Experiment~1, the agentic approach increases average latency
from 4.8s to 32.4s (+27.6s) and average token consumption from
1.8K to 27.2K ($\sim$15$\times$). The latency overhead is dominated
by LLM inference across multiple iterations rather than Jira API
round-trips, as evidenced by the modest 1.3 average tool calls per
query: most queries resolve in a single generate-execute cycle. The
naive baseline is notably cheap (1.5--2.4K tokens, 2--14s), since
it produces JQL in a single forward pass with no tool interaction.
Token consumption in the agentic setting varies substantially across
models, from roughly 17K (GPT-5.2, Pixtral Large) to 39K (GPT-OSS
120B), reflecting differences in prompt encoding efficiency and
chain-of-thought verbosity.

In Experiment~2, JiraAnchor adds exactly one tool call on average
(2.0 with compared to 1.2 without), corresponding to the single-call
retrieval design described in Section~\ref{sec:tools}. This extra
call increases average latency from 18.3s to 52.7s (+34.4s), with
the additional time spent fetching all candidate field values from
the Jira REST API and computing similarity scores. Token usage
increases modestly from 32K to 40K on average. The latency overhead
is acceptable for interactive use cases and could be reduced through
caching of field-value embeddings across queries within the same
project.

% ===========================================================================
% SECTION 6: DISCUSSION
% ===========================================================================
\section{Discussion}\label{sec:discussion}

The two experiments reveal complementary strengths of the agentic
approach. JiraAnchor delivers large, consistent gains on
field-value queries (improvement of 23.0\% overall, improvement of 49.3\% on components),
demonstrating that live value grounding is essential when
instance-specific categorical values cannot be inferred from model
training data. The agentic execution loop, by contrast, provides
selective rather than uniform improvement: gains are concentrated on
Short NL queries (improvement of 3.9\% average), where minimal context forces the
model to make assumptions that iterative execution can correct, while
Long NL queries show marginal change (improvement of 0.3\%) due to occasional
over-refinement of already-correct queries.

The overall improvement in Experiment~1 (improvement of 2.0\% across 9 models)
is variant-dependent: the agentic approach
is most valuable on under-specified queries (Short NL, Semantically
Similar) where execution feedback compensates for linguistic
ambiguity, but provides little advantage on unambiguous inputs
(Semantically Exact) and can occasionally harm performance on
well-specified queries (Long NL) through unnecessary refinement.

The error taxonomy further clarifies the boundary of what tools can
and cannot address. The dominant failure modes (Issue Type
Interpretation, 41\% of Sem.\ Similar failures; Text Field
Selection, 31\% of Short NL failures; and Version Confusion) are
semantic interpretation challenges rooted in inherent ambiguities in
the natural language or conventions in the benchmark construction.
These cannot be resolved by additional tool calls or execution
feedback; they require either user clarification or richer contextual
grounding beyond the current tool suite.

From a practical standpoint, our cost analysis shows that the
agentic approach incurs a $\sim$7$\times$ latency increase and
$\sim$15$\times$ token increase per query compared to naive
generation, while JiraAnchor adds one additional tool call with a
$\sim$3$\times$ latency overhead. For deployments where accuracy on
ambiguous queries justifies the added cost, the full agentic system
with JiraAnchor is recommended. For latency-sensitive or
cost-constrained settings, a hybrid strategy (naive generation for
well-formed queries with selective agentic fallback for ambiguous
inputs) may offer a better trade-off. The per-field regressions
observed with JiraAnchor further motivate confidence-based selective
invocation as a direction for future work.

% ===========================================================================
% LIMITATIONS
% ===========================================================================
\section{Limitations}\label{sec:limitations}

While Agentic Jackal advances text-to-JQL through live execution and semantic value grounding, several limitations remain. All experiments run against a single Jira instance, but we mitigate this by evaluating on a large-scale deployment with 22 projects, 200,000+ issues, and diverse field schemas, and by stratifying queries across clause counts, field types, and NL variants to ensure broad coverage. The agentic execution loop introduces latency and token overhead (approximately 7$\times$ and 15$\times$, respectively), but most queries resolve in a single generate-execute cycle (1.3 average tool calls), and prompt caching~\citep{lumer2026dontbreakcache} can further reduce multi-turn token costs by reusing cached system prompt and schema prefixes. JiraAnchor's embedding-based value resolution can over-match when candidate values are semantically close (e.g., ``4.7'' and ``4.7.0''), contributing to per-field regressions in Experiment~2, but the tool still delivers a net improvement of 23.0 percentage points across all 9 models, and incorporating structured matching heuristics alongside embedding similarity could improve precision on version-style values in future iterations.

% ===========================================================================
% CONCLUSION
% ===========================================================================
\section{Conclusion}\label{sec:conclusion}

We introduce Agentic Jackal, a tool-augmented multi-step agent that equips LLMs with live Jira query execution via the Jira MCP server, and JiraAnchor, a novel semantic field-value retrieval tool that resolves natural language mentions of categorical values against a live Jira instance using grep-based and embedding-based similarity search. Naive single-turn LLMs that generate queries without tool access cannot discover which categorical field values actually exist in a given instance and cannot verify or refine their output against a live data source, limiting accuracy on ambiguous or paraphrased user requests. We evaluate 9 frontier LLMs on Jackal-1K, a 1,000-query stratified subset of the Jackal benchmark, across two execution accuracy experiments. JiraAnchor improves average field-value query accuracy from 48.7\% to 71.7\% across 9 models and three categorical fields, with component accuracy rising from 16.9\% to 66.2\%. The full agentic approach improves seven of nine models, with gains concentrated on the most linguistically challenging query variants, including a 9.0\% relative improvement on Short NL queries. Manual error analysis reveals that semantic interpretation ambiguities (issue type disambiguation, text field selection, and version confusion) account for 58\% of Short NL and 68\% of Semantically Similar failures, indicating that dominant errors stem from inherent natural language ambiguity rather than value resolution failures. These findings motivate future research into intent-level disambiguation through user clarification mechanisms and prompt-cache strategies, complementing continued improvements in enterprise Jira query generation.

\clearpage

% ===========================================================================
% BIBLIOGRAPHY
% ===========================================================================
\bibliographystyle{icml2025}
\bibliography{references}

@article{frank2025jackal,
  title={Jackal: A Real-World Execution-Based Benchmark Evaluating Large Language Models on Text-to-JQL Tasks},
  author={Frank, Kevin and Gulati, Anmol and Lumer, Elias and Campagna, Sindy and Subbiah, Vamse Kumar},
  journal={arXiv preprint arXiv:2509.23579},
  year={2025}
}

@misc{atlassian2025jql,
  title={Advanced Searching Using {JQL}},
  author={{Atlassian}},
  year={2025},
  howpublished={\url{https://support.atlassian.com/jira-service-management-cloud/docs/use-advanced-search-with-jql/}},
  note={Accessed: 2025}
}

@misc{atlassian2025intelligence,
  title={Atlassian Intelligence},
  author={{Atlassian}},
  year={2025},
  howpublished={\url{https://www.atlassian.com/platform/artificial-intelligence}},
  note={Accessed: 2025}
}

@misc{atlassian2025mcp,
  title={{MCP} Server for {Jira}},
  author={{Atlassian}},
  year={2025},
  howpublished={\url{https://github.com/atlassian/mcp-server-jira}},
  note={Accessed: 2025}
}

@misc{clovity2025,
  title={{JQL} {AI} Assistant for {Jira}},
  author={{Clovity}},
  year={2025},
  howpublished={\url{https://marketplace.atlassian.com/}},
  note={Atlassian Marketplace. Accessed: 2025}
}

@misc{zhong2017seq2sql,
  title={Seq2{SQL}: Generating Structured Queries from Natural Language using Reinforcement Learning},
  author={Zhong, Victor and Xiong, Caiming and Socher, Richard},
  year={2017},
  eprint={1709.00103},
  archivePrefix={arXiv},
  primaryClass={cs.CL}
}

@inproceedings{yu-etal-2018-spider,
  title={Spider: A Large-Scale Human-Labeled Dataset for Complex and Cross-Domain Semantic Parsing and Text-to-{SQL} Task},
  author={Yu, Tao and Zhang, Rui and Yang, Kai and Yasunaga, Michihiro and Wang, Dongxu and Li, Zifan and Ma, James and Li, Irene and Yao, Qingning and Roman, Shanelle and others},
  booktitle={Proceedings of the 2018 Conference on Empirical Methods in Natural Language Processing},
  pages={3911--3921},
  year={2018}
}

@inproceedings{li-etal-2023-bird,
  title={Can {LLM} Already Serve as A Database Interface? A Big Bench for Large-Scale Database Grounded Text-to-{SQL}s},
  author={Li, Jinyang and Hui, Binyuan and Qu, Ge and Yang, Jiaxi and Li, Binhua and Li, Bowen and Wang, Bailin and Qin, Bowen and Geng, Ruiying and Huo, Nan and others},
  booktitle={Advances in Neural Information Processing Systems},
  year={2024}
}

@article{lei2025spider,
  title={Spider 2.0: Evaluating Language Models on Real-World Enterprise Text-to-{SQL} Workflows},
  author={Lei, Fangyu and Chen, Jixuan and Ye, Yuxiao and Cao, Ruisheng and Shin, Dongchan and Su, Hongjin and Suo, Zhaoqing and Gao, Hongcheng and Hu, Wenjing and Yin, Pengcheng and others},
  journal={arXiv preprint arXiv:2411.07763},
  year={2025}
}

@inproceedings{nan2026diver,
  title={{DIVER}: A Robust Text-to-{SQL} System with Dynamic Interactive Value Linking and Evidence Reasoning},
  author={Nan, Yafeng and Sun, Haifeng and Zhuang, Zirui and Qi, Qi and Chu, Guojun and Liao, Jianxin and Pei, Dan and Wang, Jingyu},
  booktitle={Proceedings of the 2026 International Conference on Management of Data (SIGMOD)},
  year={2026},
  doi={10.1145/3786640}
}

@InProceedings{10.1007/978-3-032-15632-7_2,
author="Lumer, Elias
and Gulati, Anmol
and Subbiah, Vamse Kumar
and Basavaraju, Pradeep Honaganahalli
and Burke, James A.",
editor="Marcelloni, Francesco
and Madani, Kurosh
and van Stein, Niki
and Filipe, Joaquim",
title="ScaleMCP: Dynamic and Auto-synchronizing Model Context Protocol Tools for LLM Agents",
booktitle="Computational Intelligence",
year="2026",
publisher="Springer Nature Switzerland",
address="Cham",
pages="23--42",
isbn="978-3-032-15632-7"
}

@article{lumer2024toolshed,
  title={Toolshed: Scale tool-equipped agents with advanced rag-tool fusion and tool knowledge bases},
  author={Lumer, Elias and Subbiah, Vamse Kumar and Burke, James A and Basavaraju, Pradeep Honaganahalli and Huber, Austin},
  journal={arXiv preprint arXiv:2410.14594},
  year={2024}
}

@article{lumer2025memtool,
  title={Memtool: Optimizing short-term memory management for dynamic tool calling in llm agent multi-turn conversations},
  author={Lumer, Elias and Gulati, Anmol and Subbiah, Vamse Kumar and Basavaraju, Pradeep Honaganahalli and Burke, James A},
  journal={arXiv preprint arXiv:2507.21428},
  year={2025}
}

@article{lumer2025tool,
  title={Tool and Agent Selection for Large Language Model Agents in Production: A Survey},
  author={Lumer, Elias and Gulati, Anmol and Nizar, Faheem and Hedroits, Dzmitry and Mehta, Atharva and Hwangbo, Henry and Subbiah, Vamse Kumar and Basavaraju, Pradeep Honaganahalli and Burke, James A},
  year={2025},
  publisher={Preprints}
}

@article{lumer2025tooltoagent,
  title={Tool-to-agent retrieval: Bridging tools and agents for scalable llm multi-agent systems},
  author={Lumer, Elias and Nizar, Faheem and Gulati, Anmol and Basavaraju, Pradeep Honaganahalli and Subbiah, Vamse Kumar},
  journal={arXiv preprint arXiv:2511.01854},
  year={2025}
}

@article{nizar2025agent,
  title={Agent-as-a-Graph: Knowledge Graph-Based Tool and Agent Retrieval for LLM Multi-Agent Systems},
  author={Nizar, Faheem and Lumer, Elias and Gulati, Anmol and Basavaraju, Pradeep Honaganahalli and Subbiah, Vamse Kumar},
  journal={arXiv preprint arXiv:2511.18194},
  year={2025}
}

@inproceedings{patil2025bfcl,
  author    = {Shishir G. Patil and Huanzhi Mao and Fanjia Yan and Charlie Cheng{-}Jie Ji and Vishnu Suresh and Ion Stoica and Joseph E. Gonzalez},
  title     = {The Berkeley Function Calling Leaderboard (BFCL): From Tool Use to Agentic Evaluation of Large Language Models},
  booktitle = {Proceedings of the 42nd International Conference on Machine Learning (ICML)},
  year      = {2025},
  url       = {https://openreview.net/forum?id=2GmDdhBdDk}
}

@misc{heule2025_semantic_search_agents,
  title        = {Improving agent with semantic search},
  author       = {Heule, Stefan and Jia, Emily and Jain, Naman},
  year         = {2025},
  howpublished = {\url{https://cursor.com/blog/semsearch}},
  note         = {Cursor research blog post}
}

@misc{huang2025_filesystem_context_engineering,
  title        = {How agents can use filesystems for context engineering},
  author       = {Huang, Nick},
  year         = {2025},
  howpublished = {LangChain Blog},
  url          = {http://bit.ly/41suMqx},
  note         = {Accessed: 2025}
}

@article{lumer2025rethinking,
  title={Rethinking Retrieval: From Traditional Retrieval Augmented Generation to Agentic and Non-Vector Reasoning Systems in the Financial Domain for Large Language Models},
  author={Lumer, Elias and Melich, Matt and Zino, Olivia and Kim, Elena and Dieter, Sara and Basavaraju, Pradeep Honaganahalli and Subbiah, Vamse Kumar and Burke, James A and Hernandez, Roberto},
  journal={arXiv preprint arXiv:2511.18177},
  year={2025}
}

@misc{yao2023react,
  title={{ReAct}: Synergizing Reasoning and Acting in Language Models},
  author={Yao, Shunyu and Zhao, Jeffrey and Yu, Dian and Du, Nan and Shafran, Izhak and Narasimhan, Karthik and Cao, Yuan},
  year={2023},
  eprint={2210.03629},
  archivePrefix={arXiv},
  primaryClass={cs.CL}
}

@inproceedings{pourreza2023dinsql,
  title={{DIN-SQL}: Decomposed In-Context Learning of Text-to-{SQL} with Self-Correction},
  author={Pourreza, Mohammadreza and Rafiei, Davood},
  booktitle={Advances in Neural Information Processing Systems},
  year={2023}
}

@article{wang2024macsql,
  title={{MAC-SQL}: A Multi-Agent Collaborative Framework for Text-to-{SQL}},
  author={Wang, Bing and Ren, Changyu and Yang, Jian and Liang, Xinnian and Bai, Jiaqi and Chai, Linzheng and Yan, Zhao and Zhang, Qian-Wen and Yin, Di and Sun, Xing and Li, Zhoujun},
  journal={arXiv preprint arXiv:2312.11242},
  year={2024}
}

@article{talaei2024chess,
  title={{CHESS}: Contextual Harnessing for Efficient {SQL} Synthesis},
  author={Talaei, Shayan and Pourreza, Mohammadreza and Chang, Yu-Chen and Mirhoseini, Azalia and Saberi, Amin},
  journal={arXiv preprint arXiv:2405.16755},
  year={2024}
}

@article{kamoi2024selfcorrection,
  title={When Can {LLMs} Actually Correct Their Own Mistakes? A Critical Survey of Self-Correction of {LLMs}},
  author={Kamoi, Ryo and Zhang, Yusen and Zhang, Nan and Han, Jiawei and Zhang, Rui},
  journal={Transactions of the Association for Computational Linguistics},
  volume={12},
  pages={1417--1440},
  year={2024}
}

@article{wang2025enterpriselargelanguagemodel,
  title={Enterprise Large Language Model Benchmarking for Software Engineering Tasks},
  author={Wang, Yue and others},
  journal={arXiv preprint},
  year={2025}
}

@misc{lumer2026dontbreakcache,
  title={Don't Break the Cache: An Evaluation of Prompt Caching for Long-Horizon Agentic Tasks},
  author={Lumer, Elias and Nizar, Faheem and Jangiti, Akshaya and Frank, Kevin and Gulati, Anmol and Phadate, Mandar and Subbiah, Vamse Kumar},
  year={2026},
  eprint={2601.06007},
  archivePrefix={arXiv},
  primaryClass={cs.CL}
}

@misc{kulkarni2023text2jql,
  title={Text2{JQL}},
  author={Kulkarni, Shreyas},
  year={2023},
  howpublished={\url{https://huggingface.co/datasets/shreyaskulkarni/Text2JQL}},
  note={Accessed: 2025}
}

@misc{kulkarni2023text2jql_v2,
  title={Text2{JQL}\_v2},
  author={Kulkarni, Shreyas},
  year={2023},
  howpublished={\url{https://huggingface.co/datasets/shreyaskulkarni/Text2JQL_v2}},
  note={Accessed: 2025}
}

@misc{gulati2026rowsreasoningagenticretrieval,
      title={Beyond Rows to Reasoning: Agentic Retrieval for Multimodal Spreadsheet Understanding and Editing}, 
      author={Anmol Gulati and Sahil Sen and Waqar Sarguroh and Kevin Paul},
      year={2026},
      eprint={2603.06503},
      archivePrefix={arXiv},
      primaryClass={cs.CL},
      url={https://arxiv.org/abs/2603.06503}, 
}

\clearpage
\appendix
\onecolumn

%% ----------------------------------------------------------------
%% Appendix A: Full JQL Query Examples (mirrors Jackal Appendix A)
%% ----------------------------------------------------------------
\section{Full JQL Query Examples}
\label{appendix:jql_examples}

Table~\ref{tab:jql_examples_full} presents representative JQL queries at varying clause complexities, each paired with four natural language request variants. These examples are drawn from the Jackal benchmark~\citep{frank2025jackal} and illustrate the range of linguistic variation in the evaluation data.

\newcolumntype{L}[1]{>{\raggedright\arraybackslash}p{#1}}
\setlength{\LTleft}{0pt}
\setlength{\LTright}{0pt}

\begin{center}
\small
\begin{longtable}{L{0.06\textwidth} L{0.16\textwidth} L{0.22\textwidth} L{0.15\textwidth} L{0.15\textwidth} L{0.10\textwidth}}
\caption{JQL queries with increasing clause complexity, each paired with four natural language request variants.\label{tab:jql_examples_full}}\\
\toprule
\textbf{Clauses} & \textbf{JQL} & \textbf{Long NL} & \textbf{Sem.\ Exact} & \textbf{Sem.\ Similar} & \textbf{Short NL} \\
\midrule
\endfirsthead
\toprule
\textbf{Clauses} & \textbf{JQL} & \textbf{Long NL} & \textbf{Sem.\ Exact} & \textbf{Sem.\ Similar} & \textbf{Short NL} \\
\midrule
\endhead
\bottomrule
\endfoot

2 &
\ttfamily\footnotesize updated <= -90d AND issuetype in ("Epic") &
I'm checking for epics that haven't been updated in the last 90 days, so we can identify any long-standing items that might need attention or could potentially be closed. &
Updated is less than or equal to 90 days ago, and issue type is Epic &
Large-scale tasks that haven't been changed in the last three months &
Epics inactive for 90+ days
\\[0.6em]

3 &
\ttfamily\footnotesize created >= -4w AND assignee is EMPTY AND issuetype in ("User Story") &
I'm checking for any user stories that have been created in the last four weeks but haven't been assigned to anyone yet, so we can make sure nothing important is slipping through the cracks. &
Created is within the last 4 weeks, assignee is empty, and issue type is User Story &
Stories added in the last month that haven't been assigned to anyone yet &
Unassigned user stories created in last 4 weeks
\\[0.6em]

4 &
\ttfamily\footnotesize updated >= "2025-01-01" AND issuetype in ("Bug") AND priority is not EMPTY AND description \~{} "crash" &
I'm searching for bug reports that have been updated since the start of 2025, specifically those that mention a crash in their description and have a priority set. This helps me focus on recent and prioritized crash-related issues that might need urgent attention. &
Updated is on or after 2025-01-01, issue type is Bug, priority is not empty, and description contains crash &
Problems flagged as bugs, with a set urgency level, mentioning crashes, and modified after the start of 2025 &
Bugs with priority and crash in description updated since 2025
\\[0.6em]

5 &
\ttfamily\footnotesize updated <= "2025-01-01" AND description \~{} "error" AND affectedVersion is not EMPTY AND resolution is EMPTY AND issuetype in ("Epic", "User Story", "Task", "Sub-task") &
I'm looking for any Epics, User Stories, Tasks, or Sub-tasks that mention `error' in their description, have at least one affected version specified, haven't been resolved yet, and were last updated on or before January 1st, 2025. &
Updated is on or before 2025-01-01, description contains error, affected version is not empty, resolution is empty, and issue type is Epic, User Story, Task, or Sub-task &
Items that mention an error, haven't been resolved yet, are linked to a specific version, and were last changed before January 2025, covering all major and minor work categories. &
Unresolved issues with error in description
\\

\end{longtable}

\par\footnotesize\textit{Note: Each JQL query is paired with four natural language variants: Long NL (extended description with context), Semantically Exact (literal mapping to JQL), Semantically Similar (paraphrased intent), and Short NL (concise label).}

\end{center}

%% ----------------------------------------------------------------
%% Appendix B: NL Variant Generation Prompts (mirrors Jackal Appendix B)
%% ----------------------------------------------------------------
\section{NL Variant Generation Prompts}
\label{appendix:query_variations}

The Jackal benchmark~\citep{frank2025jackal} generates four natural language variants per JQL query using the prompt templates below. Each template enforces a distinct level of abstraction, from literal translation to creative paraphrase.

\subsection{Long Natural Language (Long NL)}
\begin{verbatim}
Task: Convert a JQL query into a longer natural language sentence 
or paragraph. Include context or reasoning that someone might give 
when discussing the query aloud. Vary the style, making some 
responses conversational or explanatory.

Examples:
1. JQL: project = QTBUG AND issuetype = Bug AND status = "Open"  
   NL: I'm reviewing all open bugs in the QTBUG project so we 
       can track unresolved issues before the next sprint.

2. JQL: created >= -5d AND project = PYSIDE  
   NL: I want to look at issues reported in the last 5 days in 
       the PYSIDE project to see what's newly come in.

Given this JQL: {jql}

OUTPUT FORMAT:
Only respond with the natural language. 
Do not include any additional text or explanations.
\end{verbatim}

\subsection{Short Natural Language (Short NL)}
\begin{verbatim}
Task: Convert a JQL query into a concise natural language phrase, 
just a few words. Prefer minimal and direct expressions.

Examples:
1. JQL: project = QDS AND priority = "P0: Blocker"  
   NL: QDS blockers

2. JQL: resolution = Duplicate  
   NL: Duplicate issues

Given this JQL: {jql}

OUTPUT FORMAT:
Only respond with the natural language.
Do not include any additional text or explanations.
\end{verbatim}

\subsection{Semantically Similar}
\begin{verbatim}
Task: Convert the following JQL query into a natural language 
sentence that expresses the same intent, but uses different wording. 
Do not directly reuse JQL field names or values. Instead, rephrase 
using synonyms, conversational language, or implied meaning. Be 
creative, but maintain accuracy.

Examples:
1. JQL: status = "Open"  
   NL: Tickets that are still in progress

2. JQL: resolution = Duplicate  
   NL: Issues already reported before

Given this JQL: {jql}

OUTPUT FORMAT:
Only respond with the natural language. 
Do not include any additional text or explanations.
\end{verbatim}

\subsection{Semantically Exact}
\begin{verbatim}
Task: Translate a JQL query into a natural language sentence that 
mirrors the JQL structure and wording as closely as possible. Do not 
paraphrase or add unnecessary context. Use literal conversions of 
fields and values, preserving the order and logic.

Examples:
1. JQL: project = QTBUG AND issuetype = Bug AND status = "Open"  
   NL: Project is QTBUG, issue type is Bug, and status is Open

2. JQL: priority = "P1: Critical"  
   NL: Priority is P1: Critical

Given this JQL: {jql}

OUTPUT FORMAT:
Only respond with the natural language.
Do not include any additional text or explanations.
\end{verbatim}

%% ----------------------------------------------------------------
%% Appendix C: Agentic System Prompt (replaces Jackal Appendix C)
%% ----------------------------------------------------------------
\section{Agentic System Prompt}
\label{appendix:system_prompt}

The following is an abridged version of the schema-grounded system prompt provided to the agent at every turn. The prompt encodes the complete Jira field schema, disambiguation heuristics for ambiguous natural language, and explicit instructions for tool use. Double braces (\texttt{\{\{...\}\}}) are Python \texttt{str.format} escapes that render as single braces at runtime.

\subsection{Core Instructions}
\begin{verbatim}
You are a Jira assistant. When users ask questions about Jira issues:

1. Generate valid JQL (Jira Query Language) using only fields from 
   the SCHEMA below
2. ALWAYS use `jira_search` to execute the final query - never 
   return JQL as plain text
3. Don't ask clarifying questions - make reasonable assumptions 
   when details are ambiguous
\end{verbatim}

\subsection{Field Disambiguation Rules (Excerpt)}
\begin{verbatim}
FIELD RULES (for ambiguous natural language):

Priority:
- Hierarchy: P0 (Blocker) > P1 (Critical) > P2 (Important) >
  P3 (Somewhat important) > P4 (Low) > P5 (Not important)
- Exact value names ("P4", "Critical") -> match exactly
- "high priority" / "urgent" -> typically P1 or P2
- "low priority" / "minor" -> typically P4 or P5

Resolution:
- "unresolved" / "open" / "not resolved" -> resolution IS EMPTY
- "resolved" / "closed" / "completed" -> resolution IS NOT EMPTY

Version fields:
- affectedVersion = where bug was FOUND
- fixVersion = where fix is TARGETED

Labels vs Component:
- labels = user-applied tags (e.g., tech-debt, needs-triage)
- component = code modules/subsystems (e.g., Billing, API Gateway)
\end{verbatim}

\subsection{Text Search Guidelines}
\begin{verbatim}
TEXT SEARCH (choosing between summary and description):

Step 1: Generate BOTH complete JQL options:
  - JQL A (summary): [full JQL using summary ~ "term"]
  - JQL B (description): [full JQL using description ~ "term"]

Step 2: Compare both to the user's query:
  - summary = short titles, issue names, keywords
  - description = detailed explanations, logs, technical details

Step 3: Use the better JQL in jira_search. Never use both fields
together.
\end{verbatim}

\subsection{Tool Usage Directives}
\begin{verbatim}
TOOL USAGE (JiraAnchor):

When user mentions a SPECIFIC value but exact Jira name is unclear:
1. Call JiraAnchor to find the exact value name
2. Then call jira_search with that exact value in JQL

CALL JiraAnchor when:
- Version numbers: "5.0", "3.2.1" -> search fixVersion/affectedVersion
- Technical terms that could be components: "Networking", "Payments"
- Tag-like terms: "tech-debt", "flaky-test" -> search labels

DO NOT use for existence checks ("issues without labels"
  -> labels IS EMPTY)
\end{verbatim}

\subsection{Embedded Schema (Abridged)}
The full schema enumerates 15 Jira fields with their JQL keys, types, operators, and allowed values. Fields with large or dynamic value sets (\texttt{fixVersion}, \texttt{affectedVersion}, \texttt{component}, \texttt{labels}) include a directive to call JiraAnchor rather than guessing. Appendix~\ref{appendix:jql_schema} provides the complete field listing.

%% ----------------------------------------------------------------
%% Appendix D: Summarized JQL Schema (mirrors Jackal Appendix D)
%% ----------------------------------------------------------------
\section{Summarized JQL Schema}
\label{appendix:jql_schema}

This appendix provides a summarized version of the Jira schema used to ground JQL generation. It lists the major fields, their JQL keys, supported operator types, and representative value categories. All project-specific or proprietary values have been omitted.

\begin{itemize}
    \item \textbf{Issue Type (issuetype)} -- categorical; operators: \texttt{=, !=, IN, NOT IN}; example values: Bug, Epic, User Story, Task, Sub-task.
    \item \textbf{Project (project)} -- categorical; operators: \texttt{=, !=, IN, NOT IN}; 22 projects including QTBUG, QDS, PYSIDE.
    \item \textbf{Components (component)} -- categorical; operators: \texttt{=, !=, IN, NOT IN, IS EMPTY, IS NOT EMPTY}; large enum set (214+ values, not listed).
    \item \textbf{Platforms (custom field)} -- categorical; operators: \texttt{=, !=, IN, NOT IN, IS EMPTY, IS NOT EMPTY}; example values: Windows, Linux, macOS, Android, iOS.
    \item \textbf{Labels (labels)} -- categorical; operators: \texttt{=, !=, IN, NOT IN, IS EMPTY, IS NOT EMPTY}; values drawn from natural language mentions.
    \item \textbf{Fix Version/s (fixVersion)} -- categorical; operators: same as Labels; values omitted due to size (397+ per project).
    \item \textbf{Affects Version/s (affectedVersion)} -- categorical; operators: same as Labels; values omitted due to size.
    \item \textbf{Resolution (resolution)} -- categorical; operators: \texttt{=, !=, IN, NOT IN, IS EMPTY, IS NOT EMPTY}; example values: Fixed, Duplicate, Invalid, Won't Do.
    \item \textbf{Priority (priority)} -- categorical; operators: \texttt{=, !=, IN, NOT IN}; example values: Blocker, Critical, Important, Low.
    \item \textbf{Summary (summary)} -- text search; operators: \texttt{\~{}, !\~{}}.
    \item \textbf{Description (description)} -- text search; operators: \texttt{\~{}, !\~{}}.
    \item \textbf{Assignee (assignee)} -- categorical/special; operators: \texttt{=, !=, IS EMPTY, IS NOT EMPTY}.
    \item \textbf{Created (created)} -- date; operators: \texttt{>=, <=, >, <, =}; supports absolute dates (YYYY-MM-DD), relative dates (e.g., -5d, -4w), and functions (e.g., startOfMonth()).
    \item \textbf{Updated (updated)} -- date; same operators and hints as Created.
    \item \textbf{Resolved (resolutiondate)} -- date; same operators and hints as Created.
\end{itemize}

\paragraph{Aliases.} Common natural language mentions are normalized to schema keys.
For example: ``issue type'' $\rightarrow$ \texttt{issuetype},
``fix version/s'' $\rightarrow$ \texttt{fixVersion},
``affects version/s'' $\rightarrow$ \texttt{affectedVersion},
``components'' $\rightarrow$ \texttt{component},
``platforms'' $\rightarrow$ custom field.

%% ----------------------------------------------------------------
%% Appendix E: Agent Transcript Example (new for Agentic Jackal)
%% ----------------------------------------------------------------
\section{Agent Transcript Example}
\label{appendix:transcript}

Below is an annotated transcript from the Experiment~2 evaluation (Claude Opus 4.6, \texttt{component} field, with JiraAnchor enabled). It illustrates how the agent decomposes an ambiguous natural language request into parallel JiraAnchor queries, synthesizes the ranked results, and constructs a verified JQL query. Scores are cosine similarity values returned by JiraAnchor's embedding-based retrieval (see Section~\ref{sec:tools}).

\smallskip\noindent\textbf{Natural Language Input:}
\textit{``Tasks related to the QTBUG initiative that focus on string handling and Unicode support.''}

\smallskip\noindent\textbf{Gold JQL:}
\texttt{project IN ("QTBUG") AND component IN ("Core: QString and Unicode")}

\medskip\noindent\textbf{Step 1.} The agent identifies two semantic concepts, ``string handling'' and ``Unicode support,'' and issues parallel JiraAnchor queries for each against the 214 component values in project QTBUG.

\smallskip\noindent\fbox{\parbox{0.95\textwidth}{
\textbf{Agent $\rightarrow$ JiraAnchor} (parallel call 1)\\[0.2em]
\texttt{\{"field": "component", "query": "string handling", "projects": ["QTBUG"]\}}
}}

\smallskip\noindent\fbox{\parbox{0.95\textwidth}{
\textbf{Tool Response} (top 4 of 10 matches)\\[0.2em]
\begin{tabular}{@{}ll@{}}
GUI: Text handling & (0.529) \\
GUI: Font handling & (0.359) \\
Core: URL Handling & (0.348) \\
\textbf{Core: QString and Unicode} & \textbf{(0.336)} \\
\end{tabular}
}}

\smallskip\noindent\fbox{\parbox{0.95\textwidth}{
\textbf{Agent $\rightarrow$ JiraAnchor} (parallel call 2)\\[0.2em]
\texttt{\{"field": "component", "query": "Unicode support", "projects": ["QTBUG"]\}}
}}

\smallskip\noindent\fbox{\parbox{0.95\textwidth}{
\textbf{Tool Response} (top 4 of 10 matches)\\[0.2em]
\begin{tabular}{@{}ll@{}}
\textbf{Core: QString and Unicode} & \textbf{(0.490)} \\
GUI: Text handling & (0.379) \\
GUI: Font handling & (0.367) \\
GUI: Complex Input methods & (0.341) \\
\end{tabular}
}}

\medskip\noindent\textbf{Step 2.} The agent cross-references the two result sets. ``Core: QString and Unicode'' appears in both: rank 4 for ``string handling'' (0.336) and rank 1 for ``Unicode support'' (0.490). The agent selects it as the best match for the combined intent and constructs the JQL query.

\smallskip\noindent\fbox{\parbox{0.95\textwidth}{
\textbf{Agent $\rightarrow$ \texttt{jira\_search}}\\[0.2em]
\texttt{\{"jql": "project = QTBUG AND component = \textbackslash"Core: QString and Unicode\textbackslash""\}}
}}

\smallskip\noindent\fbox{\parbox{0.95\textwidth}{
\textbf{Tool Response:} Returns 10 issues. Non-empty result set confirms the query is valid.
}}

\medskip\noindent\textbf{Gold JQL:} \texttt{project IN ("QTBUG") AND component IN ("Core: QString and Unicode")}\\
\noindent\textbf{Agent JQL:} \texttt{project = QTBUG AND component = "Core: QString and Unicode"}\\
\noindent\textbf{Evaluation:} Execution-equivalent to gold query. \textbf{Correct.}

\smallskip\noindent\textit{Key observations:} (1) The correct component was not the top-ranked result for ``string handling'' (rank 4), but ranked first for ``Unicode support.'' By issuing parallel queries for both concepts, the agent identified the single component that satisfied both aspects of the request. (2) The agent verified the constructed query via \texttt{jira\_search} before returning, confirming non-empty results.

\medskip\noindent\rule{\textwidth}{0.4pt}

\smallskip\noindent\textbf{Comparison: Same query without JiraAnchor.}
When the same model (Claude Opus 4.6) processes the identical input without JiraAnchor, it cannot discover ``Core: QString and Unicode'' and must guess from context alone:

\smallskip\noindent\fbox{\parbox{0.95\textwidth}{
\textbf{Without JiraAnchor: 22 messages, 21+ tool calls}\\[0.2em]
\begin{tabular}{@{}rp{0.82\textwidth}@{}}
Steps 1--3: & Agent tries \texttt{component = "Core: String"}, \texttt{component = "Core: Unicode"}, etc. All return \textbf{0 results}. \\[0.2em]
Steps 4--7: & Agent searches field metadata, then tries text searches: \texttt{summary \~{} "string"}, \texttt{summary \~{} "unicode"}. Most return 0 results. \\[0.2em]
Steps 8--10: & Agent discovers partial matches under \texttt{component = "GUI: Text handling"} and settles on this value. \\[0.2em]
Final JQL: & \texttt{project = QTBUG AND component = "GUI: Text handling" AND (summary \~{} "string" OR summary \~{} "unicode" OR summary \~{} "QString")} \\[0.2em]
Result: & \textbf{Incorrect.} Wrong component, compensated with text filters, different result set.
\end{tabular}
}}

\smallskip\noindent Without JiraAnchor, the agent consumed 7$\times$ more tool calls, still selected the wrong component, and produced an incorrect result. With JiraAnchor, the same model resolved the correct value in 2 parallel calls and verified the query in a third.

%% Appendix F: Full Results Tables

\section{Full Experiment Results}
\label{appendix:full_results}

Table~\ref{tab:exp1_full} provides the complete per-variant breakdown for Experiment~1, and Table~\ref{tab:exp2_full} provides the per-field breakdown for Experiment~2.

\begin{table*}[t]
\centering
\small
\setlength{\tabcolsep}{5pt}
\begin{tabular}{l|cc|ccccc}
\toprule
 & \textbf{Naive} & \textbf{Agentic} & \multicolumn{5}{c}{\textbf{$\Delta$ (Agentic $-$ Naive)}} \\
\cmidrule(lr){2-2} \cmidrule(lr){3-3} \cmidrule(l){4-8}
\textbf{Model} & \makecell{\textbf{Overall}} & \makecell{\textbf{Overall}} & \makecell{\textbf{Overall}} & \makecell{\textbf{Long}\\\textbf{NL}} & \makecell{\textbf{Short}\\\textbf{NL}} & \makecell{\textbf{Sem.}\\\textbf{Exact}} & \makecell{\textbf{Sem.}\\\textbf{Similar}} \\
\midrule
Gemini 3 Flash      & .628 & \textbf{.710} & \textbf{+.082}  & \textbf{+.067}  & \textbf{+.173} & +.059           & +.028           \\
Gemini 3 Pro        & .641 & .671          & +.030           & $-$.024         & +.091           & +.012           & +.039           \\
GPT-5 Mini          & .616 & .638          & +.022           & $-$.035         & +.063           & +.020           & +.040           \\
Claude Opus 4.6     & .634 & .653          & +.019           & $-$.035         & +.047           & +.020           & \textbf{+.044}  \\
Pixtral Large       & .537 & .554          & +.017           & +.036           & $-$.008         & \textbf{+.126}  & $-$.087         \\
GPT-5.2             & \textbf{.659} & .670 & +.011           & +.040           & +.008           & $-$.004         & .000            \\
Claude Sonnet 4.5   & .633 & .644          & +.011           & +.039           & $-$.023         & +.043           & $-$.016         \\
GPT-OSS 120B        & .630 & .624          & $-$.006         & $-$.027         & +.028           & $-$.067         & \textbf{+.044}  \\
GPT-4o              & .643 & .634          & $-$.009         & $-$.032         & $-$.024         & $-$.004         & +.024           \\
\midrule
\textit{Average}    & \textit{.625} & \textit{.644} & \textit{+.020} & \textit{+.003} & \textit{+.039} & \textit{+.023} & \textit{+.013} \\
\bottomrule
\end{tabular}
\caption{Full per-variant execution accuracy, Experiment~1. $\Delta$ columns show the change (agentic $-$ naive) for each query variant. Bold marks the best value per column.}
\label{tab:exp1_full}
\end{table*}

\begin{table*}
\centering
\small
\setlength{\tabcolsep}{5pt}
\begin{tabular}{l|cc|c|cc|c|cc|c|cc|c}
\toprule
 & \multicolumn{3}{c|}{\textbf{Fix Version}} & \multicolumn{3}{c|}{\textbf{Affected Version}} & \multicolumn{3}{c|}{\textbf{Components}} & \multicolumn{3}{c}{\textbf{Overall Avg.}} \\
\cmidrule(lr){2-4} \cmidrule(lr){5-7} \cmidrule(lr){8-10} \cmidrule(l){11-13}
\textbf{Model} & \textbf{Base} & \textbf{+JA} & \textbf{$\Delta$} & \textbf{Base} & \textbf{+JA} & \textbf{$\Delta$} & \textbf{Base} & \textbf{+JA} & \textbf{$\Delta$} & \textbf{Base} & \textbf{+JA} & \textbf{$\Delta$} \\
\midrule
Claude Sonnet 4.5    & .639 & .736 & +.097           & .611 & .764 & \textbf{+.153} & .137 & .699 & +.562           & .462 & .733 & +.271           \\
Claude Opus 4.6      & .736 & \textbf{.778} & +.042  & .722 & .778 & +.056           & \textbf{.425} & .644 & +.219  & \textbf{.628} & .733 & +.105  \\
Gemini 3 Flash       & \textbf{.790} & .710 & $-$.080 & \textbf{.806} & \textbf{.871} & +.065 & .206 & .619 & +.413  & .601 & .733 & +.132           \\
Gemini 3 Pro         & .667 & .722 & +.055           & .694 & .792 & +.098           & .219 & .712 & +.493           & .527 & \textbf{.742} & +.215  \\
GPT-4o               & .569 & .736 & +.167           & .667 & .750 & +.083           & .110 & .644 & +.534           & .449 & .710 & +.261           \\
GPT-5.2              & .556 & .750 & +.194           & .583 & .736 & \textbf{+.153} & .110 & .630 & +.520           & .416 & .705 & +.289           \\
GPT-5 Mini           & .542 & .750 & \textbf{+.208} & .569 & .722 & \textbf{+.153} & .068  & .658 & +.590           & .393 & .710 & \textbf{+.317} \\
GPT-OSS 120B         & .611 & .667 & +.056           & .639 & .639 & .000             & .123 & .630 & +.507           & .458 & .645 & +.187           \\
Pixtral Large        & .569 & .722 & +.153           & .667 & .778 & +.111           & .123 & \textbf{.726} & \textbf{+.603} & .453 & \textbf{.742} & +.289 \\
\midrule
\textit{Average} & \textit{.631} & \textit{.730} & \textit{+.099} & \textit{.662} & \textit{.759} & \textit{+.097} & \textit{.169} & \textit{.662} & \textit{+.493} & \textit{.487} & \textit{.717} & \textit{+.230} \\
\bottomrule
\end{tabular}
\caption{Full per-field execution accuracy, Experiment~2 (JiraAnchor ablation). Base = Baseline Agent (without JiraAnchor), +JA = Agent with JiraAnchor. Bold marks the best value per column.}
\label{tab:exp2_full}
\end{table*}

\end{document}